\documentclass[runningheads]{llncs}
\usepackage{graphicx}

\usepackage{tikz}
\usepackage{comment}
\usepackage{amsmath,amssymb} %
\usepackage{color}

\usepackage[accsupp]{axessibility}  %

\usepackage{graphicx}
\usepackage{amsmath}
\usepackage{amssymb}
\usepackage{booktabs}
\usepackage{multirow}
\usepackage{enumitem}
\usepackage{verbatim}
\usepackage{comment}
\usepackage{bm}
\usepackage{adjustbox}
\usepackage{array}
\usepackage{pifont}
\usepackage{stmaryrd}
\usepackage{flushend}
\usepackage{makecell}
\usepackage{float}
\usepackage{xfrac}
\usepackage{rotating}
\usepackage{colortbl}
\usepackage{xspace}

\usepackage{pifont}
\usepackage{setspace}
\usepackage{cuted}
\usepackage{bm}
\usepackage{diagbox}
\usepackage[pagebackref=true,breaklinks=true,colorlinks,citecolor=blue,bookmarks=false]{hyperref}
\usepackage[capitalize]{cleveref}
\usepackage[normalem]{ulem}

\newcommand{\rotr}[1]{\rotatebox{90}{#1}}

\newcommand\Gianni{\textcolor{black}}
\newcommand\Sev{\textcolor{black}}
\newcommand\Gianninew{\textcolor{black}}
\newcommand{\Emi}[1]{#1}
\newcommand{\Xuanlong}{\textcolor{black}}

\newcommand{\ab}[1]{\textcolor{black}{#1}}

\newcommand{\comEmi}[1]{\textcolor{black}{{#1}}}

\makeatletter
\DeclareRobustCommand\onedot{\futurelet\@let@token\@onedot}
\def\@onedot{\ifx\@let@token.\else.\null\fi\xspace}

\makeatother

\newcommand{\vomega}{\boldsymbol{\mathbf{\omega}}}

\newcommand{\vX}{\mathbf{X}}

\newcommand{\vz}{\mathbf{z}}
\newcommand{\vp}{\mathbf{p}}

\newcommand{\vx}{\mathbf{x}}

\definecolor{dgreen}{rgb}{0.0, 0.5, 0.0}
\definecolor{better}{rgb}{0.19, 0.55, 0.91}
\definecolor{worse}{rgb}{0.82, 0.1, 0.26}

\newcommand{\parag}[1]{\smallskip\noindent\textbf{#1}~~}

\newcommand{\second}{\cellcolor{blue!10}}
\newcommand{\first}{\cellcolor{blue!30}}

\begin{document}
\pagestyle{headings}
\mainmatter
\def\ECCVSubNumber{1119}  %

\title{Latent Discriminant Deterministic Uncertainty} %

\titlerunning{Latent Discriminant deterministic Uncertainty}
\author{Gianni Franchi\inst{1\dagger}  \and
Xuanlong Yu\inst{1,2\dagger} \and
Andrei Bursuc\inst{3} \and
Emanuel Aldea\inst{2} \and
Severine Dubuisson\inst{4} \and
David Filliat\inst{1}}
\authorrunning{Franchi et al.}
\institute{U2IS, ENSTA Paris, Institut polytechnique de Paris \and
SATIE, Paris-Saclay University \and
valeo.ai \and
CNRS, LIS, Aix Marseille University}
\maketitle

\newcommand\blfootnote[1]{%
\begingroup
\renewcommand\thefootnote{}\footnote{#1}%
\addtocounter{footnote}{-1}%
\endgroup
}

\blfootnote{$\dagger$ Equal contribution.}

\begin{abstract}

Predictive uncertainty estimation is essential for deploying Deep Neural Networks in real-world autonomous systems. 
However, most successful approaches are computationally intensive.
In this work, we attempt to address these challenges in the context of autonomous driving perception tasks. Recently proposed Deterministic Uncertainty Methods (DUM) can only partially meet such requirements as their scalability to complex computer vision tasks is not obvious. In this work we advance a scalable and effective DUM for high-resolution semantic segmentation, that relaxes the Lipschitz constraint typically hindering practicality of such architectures.
We learn a discriminant latent space by leveraging a distinction maximization layer over an arbitrarily-sized set of trainable prototypes. Our approach achieves competitive results over  Deep Ensembles, the state of the art for uncertainty prediction,  on image classification, segmentation and monocular depth estimation tasks. Our code is available at \href{https://github.com/ENSTA-U2IS/LDU}{https://github.com/ENSTA-U2IS/LDU}.

\end{abstract}
\section{Introduction}
Uncertainty estimation and robustness are essential for deploying Deep Neural Networks (DNN) in real-world systems with different levels of autonomy, ranging from simple driving assistance functions to fully autonomous vehicles. In addition to excellent predictive performance, DNNs are also expected to address different types of uncertainty (noisy, ambiguous or out-of-distribution samples, distribution shift, etc.), while ensuring real-time computational performance.
These key and challenging requirements have stimulated numerous solutions and research directions leading to significant progress in this area~\cite{blundell2015weight,lakshminarayanan2017simple,gal2016dropout,kendall2017uncertainties,wen2020batchensemble,mukhoti2020calibrating}. Yet, the best performing approaches are computationally expensive~\cite{lakshminarayanan2017simple}, while faster variants struggle to disentangle different types of uncertainty~\cite{fort2019deep,malinin2020ensemble,mukhoti2021deterministicDDU}.

We study 
a promising new line of methods, termed deterministic uncertainty methods (DUMs)~\cite{postels2021practicality}, that has recently emerged for estimating uncertainty in a computational efficient manner from a single forward pass~\cite{mukhoti2021deterministicDDU,van2020uncertainty,van2021improvingDUE,postels2020hiddenMIR,liu2020simpleSNGP}. In order to quantify uncertainty, these methods rely on some statistical or geometrical properties of the hidden features of the DNNs. While appealing for their good 
Out-of-Distribution (OOD) uncertainty estimations at low computational cost, they have been used mainly for classification tasks and their specific regularization is often unstable when training deeper DNNs~\cite{torrnips}.
We then propose a new DUM technique, based on a discriminative latent space that improves both scalability and flexibility.
We achieve this by still following the principles of DUMs of learning a sensitive and smooth representation that mirrors well the input distribution, although not by enforcing directly the Lipschitz constraint.

Our 
DUM, dubbed Latent Discriminant deterministic Uncertainty (LDU), is based on a DNN 
imbued with a set of prototypes over its latent representations.
These prototypes act like a memory 
that allows to better analyze features from new images in light of the ``knowledge'' acquired by the DNN from the training data.
Various forms of prototypes have been studied for anomaly detection in the past~\cite{gong2019memorizing} and they often take the shape of a dictionary of representative features.
\ab{Instead, }LDU is trained to learn the optimal prototypes, such that this distance improves the accuracy and the uncertainty prediction. Indeed to train LDU, we introduce a confidence-based loss that learns to predict the error of the DNN given the data. %
ConfidNet \cite{corbiere2019addressing} and SLURP \cite{yu2021slurp} have shown that we can train an auxiliary network to predict the uncertainty, 
\ab{at the cost of a more complex training pipeline and more inference steps. }%
Here LDU is lighter, faster and 
\ab{needs only} a single forward pass. LDU can be used as a pluggable learning layer 
on top of 
DNNs. We demonstrate that LDU avoids feature collapse and can be applied to multiple computer vision tasks. In addition, LDU improves the prediction accuracy of the baseline DNN without LDU.

\parag{Contributions.} To summarize, our contributions are as follows: 
\textbf{(1)} LDU (Latent Discriminant deterministic Uncertainty): an efficient and scalable DUM approach for uncertainty quantification. \textbf{(2)} A study of LDU's properties against feature collapse. \textbf{(3)} 
Evaluations of LDU on a range of computer vision tasks and settings (image classification, semantic segmentation, depth estimation) and the implementation of a set of powerful baselines to further encourage research in this area.

\section{Related work}
In this section, we focus on the related works from two perspectives: uncertainty quantification algorithms applied to computer vision tasks and prototype learning on DNNs. In Table~\ref{tab:relatedwork}, we list various uncertainty quantification algorithms according to different computer vision tasks.

\begin{table}[t]
\begin{center}
\scalebox{0.45}
{
\begin{tabular}{lccc} 
\toprule
 & \multicolumn{3}{c}{\textbf{Computer vision tasks}} \\
\multicolumn{1}{l}{\begin{tabular}[c]{@{}c@{}}\textbf{Uncertainty estimation methods}\\\end{tabular}} & Image Classification & Semantic Segmentation & 1D/2D Regression\\ 
\toprule
Bayesian/Ensemble based methods&  \begin{tabular}[c]{@{}c@{}}Rank-1 BNN~\cite{dusenberry2020efficient},\\ PBP~\cite{hernandez2015probabilistic}, Deep Ensembles~\cite{lakshminarayanan2017simple},\\Bayes by Backprop~\cite{blundell2015weight}, MultiSWAG~\cite{wilson2020bayesian}\end{tabular} & \begin{tabular}[c]{@{}c@{}}Deep Ensembles~\cite{lakshminarayanan2017simple},\\Bayes by Backprop~\cite{blundell2015weight},\\MultiSWAG~\cite{wilson2020bayesian}\end{tabular} &  \begin{tabular}[c]{@{}c@{}}FlowNetH~\cite{ilg2018uncertainty}, PBP~\cite{hernandez2015probabilistic},\\Deep Ensembles~\cite{lakshminarayanan2017simple},\\Bayes by Backprop~\cite{blundell2015weight}, MultiSWAG~\cite{wilson2020bayesian}\end{tabular}\\ 
\midrule
Dropout/Sampling based methods& MC-Dropout\cite{gal2016dropout} & \begin{tabular}[c]{@{}c@{}}MC-Dropout\cite{gal2016dropout},\\Bayesian SegNet~\cite{kendall2015bayesian}\end{tabular} & \begin{tabular}[c]{@{}c@{}}Infer-perturbations~\cite{mi2019training},\\MC-Dropout\cite{gal2016dropout}\end{tabular} \\ 
\midrule
Learning distribution/Auxiliary network &
\begin{tabular}[c]{@{}c@{}}ConfidNet~\cite{corbiere2019addressing},\\Kendall et al.~\cite{kendall2017uncertainties}\end{tabular} & \begin{tabular}[c]{@{}c@{}}ConfidNet~\cite{corbiere2019addressing},\\Kendall et al.~\cite{kendall2017uncertainties}\end{tabular}& \begin{tabular}[c]{@{}c@{}}Hu et al.~\cite{hu2021learning}, SLURP~\cite{yu2021slurp},\\ Mono-Uncertainty~\cite{poggi2020uncertainty}, Asai et al.~\cite{asai2019multi},\\Kendall et al.~\cite{kendall2017uncertainties}, Nix et al.~\cite{nix1994uncertainty}\end{tabular}\\ 
\midrule
Deterministic uncertainty methods &
\begin{tabular}[c]{@{}c@{}}SNGP~\cite{liu2020simpleSNGP}, VIB~\cite{alemi2018uncertainty}, DUM~\cite{wu2020simple},\\DUE~\cite{van2021improvingDUE}, DDU~\cite{mukhoti2021deterministicDDU}, MIR~\cite{postels2020hiddenMIR}, DUQ~\cite{van2020uncertainty} \end{tabular}& MIR~\cite{postels2020hiddenMIR} & DUE~\cite{van2021improvingDUE}, MIR~\cite{postels2020hiddenMIR}\\
\bottomrule
\end{tabular}
}
\end{center}
\caption{{Summary of the uncertainty estimation methods applied to the specific computer vision tasks.}}
\vspace{-2em}
\label{tab:relatedwork}
\end{table}

\subsection{Uncertainty estimation for computer vision tasks}
\parag{\textbf{Uncertainty for image classification and semantic segmentation.}}
Quantifying uncertainty for classification and semantic segmentation can be done with Bayesian Neural Networks (BNNs)~\cite{blundell2015weight,hernandez2015probabilistic,wilson2020bayesian,dusenberry2020efficient}, which estimate the posterior distribution of the DNN weights to marginalize the likelihood distribution at inference time. These approaches achieve good performances on image classification, but they do not scale well to semantic segmentation.
Deep Ensembles~\cite{lakshminarayanan2017simple} achieve state-of-the-art performance on various tasks. Yet, this approach is computationally costly \ab{in both training and inference}. 
\ab{Some techniques learn a confidence score as uncertainty~\cite{corbiere2019addressing}, but struggle without sufficient negative samples to learn from.}
\ab{MC-Dropout~\cite{gal2016dropout} is a generic and easy to deploy approach, however its uncertainty is not always reliable~\cite{ovadia2019can} while requiring multiple forward passes.}
Deterministic Uncertainty Methods (DUMs) ~\cite{liu2020simpleSNGP,wu2020simple,alemi2018uncertainty,van2021improvingDUE,mukhoti2021deterministicDDU,postels2020hiddenMIR,van2020uncertainty} are new strategies that allow to quantify epistemic uncertainty in the DNNs with a single forward pass. Yet, except for MIR \cite{postels2020hiddenMIR}, to the best of our knowledge none of these techniques work on semantic segmentation.

\parag{\textbf{\\Uncertainty for 1D/2D regression.}}
Regression in computer vision comprises monocular depth estimation~\cite{bhat2021adabins,lee2019big}, optical flow estimation~\cite{teed2020raft,sun2018pwc}, or pose estimation~\cite{openpose,rogez2019lcr}.
One solution for quantifying the uncertainty consists in 
\ab{formalizing the output of a DNN as a parametric distribution and training the DNN  to estimate its parameters~\cite{nix1994uncertainty,kendall2017uncertainties}.}
Multi-hypothesis DNNs~\cite{ilg2018uncertainty} consider that the output is a Gaussian distribution and focus on optical flow. Some techniques estimate a confidence score for regression thanks to an auxiliary DNN~\cite{yu2021slurp,poggi2020uncertainty}. 
\Xuanlong{
Deep Ensembles~\cite{lakshminarayanan2017simple} for regression, consider that each DNN outputs the parameters of a Gaussian distribution, to form a mixture of Gaussian distributions.
Sampling-based methods~\cite{gal2016dropout,mi2019training} simply apply dropout or perturbations to some layers during test time to quantify the uncertainty. Yet, their computational cost
\ab{remains} important compared to a single \ab{forward} pass in the network.}
Some DUMs~\cite{van2021improvingDUE,postels2020hiddenMIR} also work on regression tasks. DUE~\cite{van2021improvingDUE} is applied in a 1D regression task and MIR~\cite{postels2020hiddenMIR} in monocular depth estimation.

\subsection{Prototype learning in DNNs}
Prototype-based learning approaches have been introduced on traditional handcrafted features~\cite{liu2001evaluation}, and have been recently applied to DNNs as well, for more robust predictions~\cite{wen2016discriminative,yang2018robust,gong2019memorizing,chen2020learning}. The center loss~\cite{wen2016discriminative} can help DNNs to build more discriminative features by compacting intra-class features and dispersing the inter-class ones. Based on this principle, Convolutional Prototype Learning (CPL)~\cite{yang2018robust} with prototype loss also improves the intra-class compactness of the latent features. 
Chen et al.~\cite{chen2020learning} try to bound the unknown classes by learning reciprocal points for better open
set recognition. Similar to~\cite{van2017neural,razavi2019generating}, MemAE~\cite{gong2019memorizing} learns a memory slot of the prototypes to strengthen the reconstruction error of anomalies in the process of the reconstruction. These prototype-based methods are well suited for classification tasks but are rarely used in semantic segmentation and regression tasks.

\section{Latent Discriminant deterministic Uncertainty (LDU)}
\label{sec:LDU}

\subsection{DUM preliminaries}

DUMs arise as a promising line of research for estimating epistemic uncertainty in conventional DNNs in a computationally efficient manner and from a single forward pass. DUM approaches generally focus on learning useful and informative hidden representations of a model~\cite{liu2020simpleSNGP,wu2020simple,alemi2018uncertainty,van2021improvingDUE,mukhoti2021deterministicDDU,postels2020hiddenMIR} by considering %
that the distribution of the hidden representation should be representative for the input distribution. Most of the conventional models suffer from the \emph{feature collapse} problem~\cite{van2020uncertainty} when OOD samples are mapped to similar feature representations as in-distribution ones, thus hindering OOD detection from these representations. DUMs address this issue through various regularization strategies for constraining the hidden representations to mimic distances from the input space. In practice this amounts to striking a balance between \emph{sensitivity} (when the input changes, the feature representation should also change) and \emph{smoothness} (a small change in the input cannot generate major shifts in the feature representation) of the model. To this end, most methods enforce constraints over the Lipschitz constant of the DNN~\cite{mandelbaum2017distance,van2020uncertainty,liu2020simpleSNGP}.

Formally, we define $f_{\omega}(\cdot)$ a DNN with trainable parameters $\vomega$, and an input sample $\vx$ from a set of images $\mathcal{X}$.
Our DNN $f_{\vomega}$ is composed of two main blocks: a feature extractor $h_{\vomega}$ and a head $g_{\vomega}$, such that $f_{\vomega}(\vx)=(g_{\vomega}\circ h_{\vomega}) (\vx)$. $h_{\vomega}(\vx)$ computes a latent representation from $\vx$, while $g_{\vomega}$ is the final layer, that takes $h_{\vomega} (\vx)$ as input, and outputs the logits of $\vx$. The bi-Lipschitz condition implies that for any pair of inputs $\vx_1$ and $\vx_2$ from $\mathcal{X}$:

\vspace{-2mm}
\begin{equation}
    L_1 \| \vx_1 -\vx_2\| \leq  \| h_{\vomega}(\vx_1) - h_{\vomega}(\vx_2)\| \leq  L_2 \| \vx_1 -\vx_2\|
\end{equation}
where $L_1$ and $L_2$ are positive and bounded Lipschitz constants $0<L_1<1<L_2$. The upper Lipschitz bound enforces the smoothness and is an important condition for the robustness of a DNN by preventing over-sensitivity to perturbations in the input space of $\vx$, i.e., the pixel space. The lower Lipschitz bound deals with the sensitivity and strives to preserve distances in the latent space as mappings of distances from the input space, i.e., preventing representations from being too smooth, thus avoiding feature collapse. 
Liu et al.~\cite{liu2020simpleSNGP} argue that for residual DNNs~\cite{he2016deep}, we can ensure $f_{\vomega}$ %
to be bi-Lipschitz by forcing its  residuals to be Lipschitz and choosing sub-unitary Lipschitz constants.

There are different approaches for imposing the bi-Lipschitz constraint over a DNN, out of which we describe the most commonly used ones in recent works~\cite{arjovsky2017wasserstein,gulrajani2017improved,miyato2018spectral,behrmann2019invertible}. %
Wasserstein GAN~\cite{arjovsky2017wasserstein} 
enforces the Lipschitz constraint by clipping the weights. However, this turns out to be prone to either vanishing or exploding gradients if the clipping threshold is  not carefully tuned~\cite{gulrajani2017improved}. An alternative solution from GAN optimization is gradient penalty~\cite{gulrajani2017improved} which is practically an additional loss term that regularizes the $L_2$ norm of the Jacobian of weight matrices of the DNN. However this can also lead to high instabilities~\cite{liu2020simpleSNGP,mukhoti2021deterministicDDU} and slower training~\cite{mukhoti2021deterministicDDU}. Spectral Normalization~\cite{miyato2018spectral,behrmann2019invertible} brings better stability and training speed, however, on the downside, it supports only a fixed pre-defined size for the input,
in the same manner as fully connected layers. For computer vision tasks, such as semantic segmentation which is typically performed on high resolution images, 
\ab{constraining} the input size is a strong limitation. 
Moreover, Postels et al.~\cite{postels2020hiddenMIR} argue that in addition to the architectural constraints, these strategies for avoiding feature collapse risk overfitting epistemic uncertainty to the task of OOD detection. This motivates us to seek a new DUM strategy that does not need the network to comply with the Lipschitz constraint. The recent MIR approach~\cite{postels2020hiddenMIR} 
\ab{advances }an alternative regularization strategy that adds a decoder branch to the network, thus forcing the intermediate activations to better cover and represent the input space. However in the case of high resolution images, reconstruction can be a challenging task and the networks can over-focus \ab{on potentially useless and uninformative} details at the cost of loss of global information. We detail our strategy below.

\subsection{Discriminant Latent space}
\Emi{An informative latent representation should project similar data samples close and dissimilar ones far away. Yet, it has long been known that in high-dimensional spaces the Euclidean distance and other related p-norms are a very poor indicator of sample similarity as most samples are nearly equally far/close to each other~\cite{aggarwal2001surprising,assent2012clustering}. At the same time, the samples of interest are often not uniformly distributed, and may be projected by means of a learned transform on a lower-dimensional manifold, namely the latent representation space.} 

Instead of focusing on preserving the potentially uninformative distance in the input space, we can rather attempt to better deal with distances in the lower-dimensional latent space. To this end, we propose to use a distinction maximization (DM) layer~\cite{macedo2021entropic} that has been recently considered as a replacement for the last layer to produce better uncertainty estimates, in particular for OOD detection~\cite{padhy2020revisiting,macedo2021entropic}. In a DM layer, the units of the classification layer are seen as representative class prototypes and the classification prediction is computed by analyzing the localization of the input sample w.r.t. all class prototypes as indicated by the negative Euclidean distance. 
Note that a similar idea has been considered in the few-shot learning literature, where DM layers are known as cosine classifiers~\cite{gidaris2018dynamic,qi2018low,snell2017prototypical}.
In contrast to all these approaches that use DM as a last layer for classification predictions, we employ it as hidden layer over latent representations.
More specifically, we insert DM in the pre-logit layer. We argue that this allows us to better guide learning and preserve the discriminative properties of the latent representations compared to placing DM as last layers where the weights are more specialized for classification decision than for feature representation. 
We can easily integrate this layer in the architecture without impacting the training pipeline. 

Formally, we denote $\vz \in \mathbb{R}^n$ the latent representation of dimension $n$ of $\vx$, i.e., $\vz{=}h_{\vomega}(\vx)$, that is given as input to the DM layer. Given a set $\vp_{\vomega}{=}\{\vp_i\}_{i=1}^m$, of $m$ vectors ( $\vp_i {\in} \mathbb{R}^n$) 
that are trainable, we define the DM layer as follows:

\begin{equation}
\mbox{DM}_{p}(\vz) =\begin{bmatrix} -\| \vz -\vp_1\|, \ldots, -\| \vz -\vp_m\| \end{bmatrix}^\top
\end{equation}

The $L_2$ distance considered in the DM layer is not bounded, thus when DM is used as intermediate layer, relying on the $L_2$ distance could cause instability during training. %
In our proposed approach,
we use instead the cosine similarity, $S_c(\cdot, \cdot)$. Our DM layer reads now:

\begin{equation}
    \mbox{DM}_{p}(\vz) =\begin{bmatrix} S_c(\vz ,\vp_1), \ldots,  S_c(\vz ,\vp_m) \end{bmatrix}^\top
\end{equation}

The vectors %
$\vp_i$ can be seen as a set of prototypes in the latent space that can help in better placing an input sample in the learned representation space using these prototypes as references. This is in contrast to prior works with DM being considered as last layer, where the prototypes represent canonical representations for samples belonging to a class~\cite{snell2017prototypical,macedo2021entropic}. Since hidden layers are used here, we can afford to consider an arbitrary number of prototypes that can define 
richer latent mapping through a finer coverage of the representation space. DM layers learn the set of weights $\{\vp_i\}_{i=1}^m$ such that the cosine similarity (evaluated between $\vz$  and the prototypes) is optimal for a given task. %

\begin{figure}
\setlength{\abovecaptionskip}{0.cm}
    \centering{\includegraphics[width=0.93\linewidth]{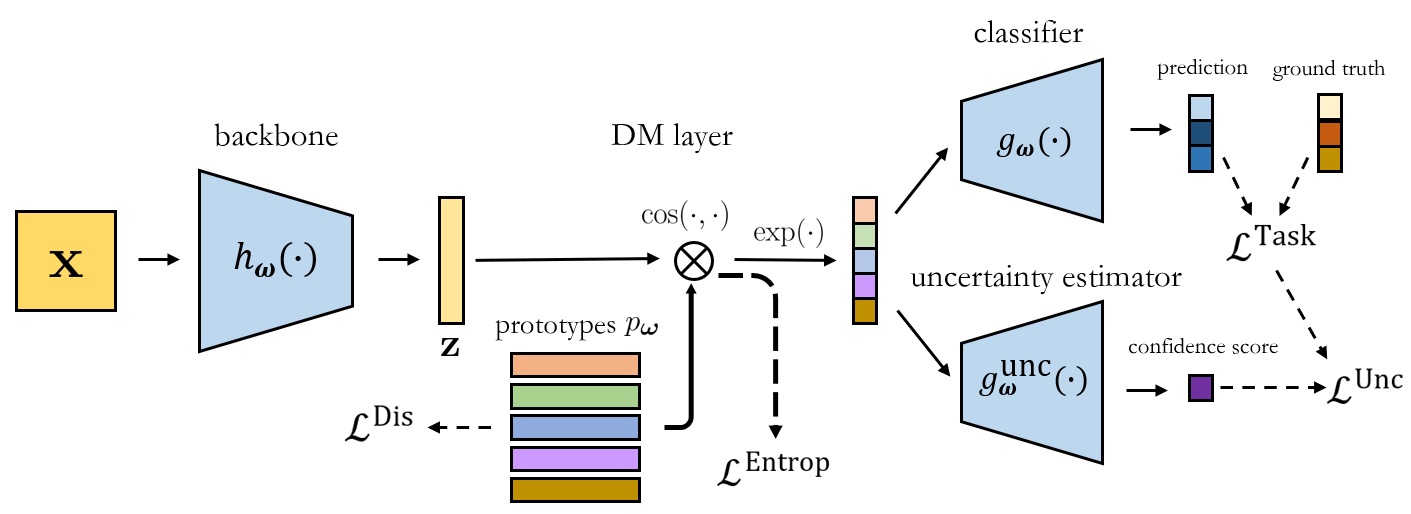}}
    \caption{ \textbf{Overview of LDU:} the DNN learns a discriminative latent space thanks to learnable prototypes $\vp_{\vomega}$. The DNN backbone computes a feature vector $\vz$ for an input $\vx$ and then the DM layer matches it with the prototypes. The computed similarities reflecting the position of $\vz$ in the learned feature space, are subsequently processed by the classification layer and the uncertainty estimation layer. The dashed arrows point to the loss functions that need to be optimized for training LDU.
    }
    \label{fig:process}
    \vspace{-3mm}
\end{figure}

We apply the distinction maximization %
on this hidden representation, and subsequently use the exponential function as 
activation function. We consider the exponential function as it can sharpen similarity values and thus facilitates the alignment of the data embedding to the corresponding prototypes in the latent space. Finally, we apply a last fully connected layer for classification on this embedding. Our DNN (see Figure \ref{fig:process}) can be written as:
\begin{equation}
    f_{\vomega}(\vx)=\left[ g_{\vomega}\circ (\mbox{exp}(-\mbox{DM}_{p}(h_{\vomega}))) \right] (\vx)  
\label{eq:full-dnn}
\end{equation}

We can see 
\ab{from} Eq.~\eqref{eq:full-dnn} that the vector weights $\vp_i$ are optimized 
jointly with the other 
DNN parameters. We argue that $\vp_i$ 
can work as indicators for analyzing and emphasizing patterns in the latent representation prior to making a classification prediction in the final layers.

\subsection{LDU optimization}\label{sec:Avoiding}

Given a DNN $f_{\omega}$ we usually optimize its parameters to minimize a loss $\mathcal{L}^{\mbox{Task}}$. 
This can lead to prototypes \ab{specialized for solving that task} that
\ab{ do} not encapsulate uncertainty relevant properties.
Hence we propose to enforce the prototypes to be linked to uncertainty first by avoiding the collapse of all prototypes to a single prototype. Second, we constrain the latent representation $\mbox{DM}_{p}(h_{\vomega})$ of the DNN to not rely 
only on a single prototype. %
Finally, we optimize an MLP $g_{\vomega}^{\mbox{unc}}$ on the top of the latent representation $\mbox{DM}_{p}(h_{\vomega})$  %
such that the output of this MLP 
provides more meaningful information for uncertainty estimation.

First, we add a loss to force the prototypes to be dissimilar: %
$$\mathcal{L}^{\mbox{Dis}} = -\sum_{i<j} \| \vp_i - \vp_j\|.$$
Then, we also add one loss to 
\ab{constrain the} latent representation to stay close to different prototypes.
We achieve this with an entropy-like loss:
$$
\mathcal{L}^{\mbox{Entrop}} = \sum_{i=1}^n \sigma(\mbox{DM}_{p}(h_{\vomega}))_i\Xuanlong{\cdot} \log(\sigma(\mbox{DM}_{p}(h_{\vomega}))_i),
$$ 
where $\sigma$ is the softmax layer, and the 
\ab{subscript} index $i$ corresponds to the $i$-th coefficient of a tensor. Different from 
per-class prototypes~\cite{wen2016discriminative,yang2018robust,chen2020learning}, we obtain more discriminative features by increasing the distance between prototypes and enlarging the dispersion of 
features corresponding to different prototypes.

We propose to train $g_{\vomega}^{\mbox{unc}}$ to predict the error of the DNN, which helps us relate the prototypes to the uncertainty. 
\ab{Formally,} given an input data $\vx$, %
 its groundtruth $y$ ($y$ can be a scalar or a vector if we deal with regression) %
 and, its loss $\mathcal{L}^{\mbox{Task}}(g_{\vomega}(\vx),y)$, we train
$g_{\omega}^{\mbox{unc}} $ by minimizing: 
$$\mathcal{L}^{\mbox{Unc}} =\mbox{BCE}(\left[ g_{\vomega}^{\mbox{unc}} \circ (\mbox{exp}(-\mbox{DM}_{p}(h_{\vomega}))) \right] (\vx), \mathcal{L}^{\mbox{Task}}(g_{\vomega}(\vx),y)),$$ 
after 
normalizing $\mathcal{L}^{\mbox{Task}}(g_{\vomega}(\vx),y)$ 
over the \ab{mini-}batch such that its maximum value is equal to one and its minimum is equal to zero. BCE stands for the binary cross entropy, which was empirically validated to perform better than common alternatives such as the mean square error and the absolute error.

{All these losses combined allow us to have a DNN which can predict uncertainty, avoid feature collapse and have the potential to improve the accuracy of the prediction. To summarize, the following loss function $\mathcal{L}^{\mbox{total}}$ will be optimized to train a DNN containing a DM layer:
\begin{equation}
    \mathcal{L}^{\mbox{total}} = \mathcal{L}^{\mbox{Task}}
    + \mathbf{\lambda}(\mathcal{L}^{\mbox{Entrop}} + \mathcal{L}^{\mbox{Dis}} + \mathcal{L}^{\mbox{Unc}})
\end{equation}
where $\mathbf{\lambda}$ is a hyper-parameter for the auxiliary losses.}

\subsection{Addressing feature collapse}
\begin{figure}[t]
\centering
\begin{tabular}{cc}
\textbf{\tiny{PCA of MLP}} & \textbf{\tiny{PCA of DM-MLP}}\\
\includegraphics[width=0.25\linewidth]{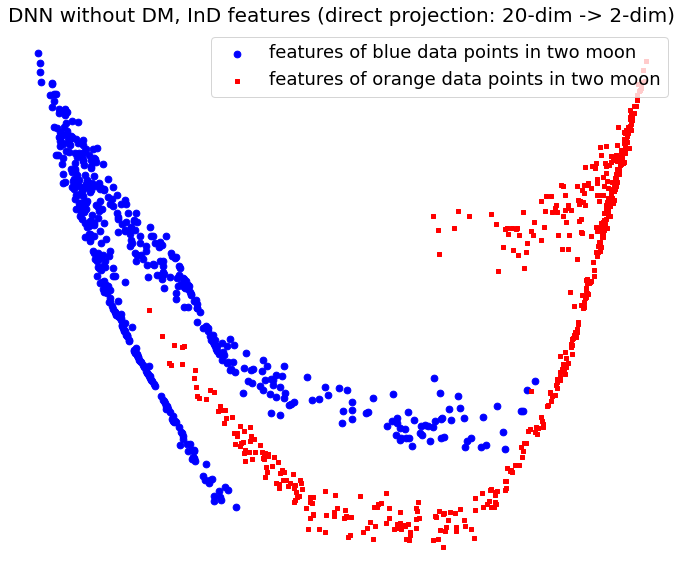}&
\includegraphics[width=0.25\linewidth]{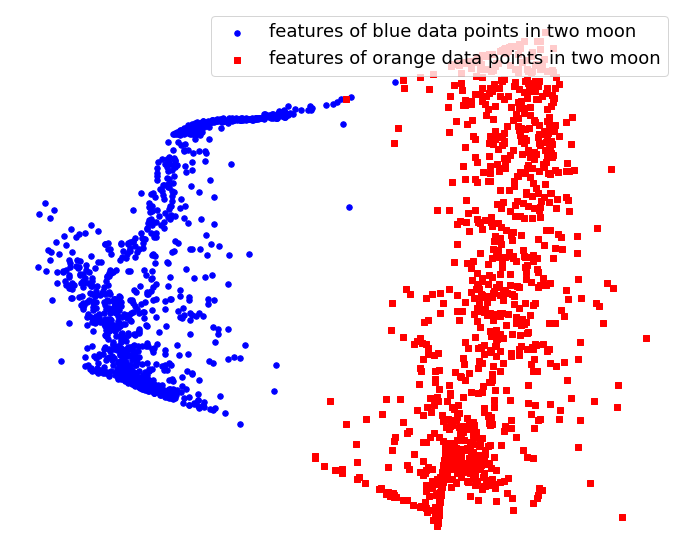}
\end{tabular}
\caption{ PCA 2D projection on the left of a standard MLP and on the right of a DM-MLP  trained on the two moons dataset. Blue and red points indicate the features of data points of the two classes respectively. As we can see, the representations on the MLP are overlapping between the two classes, leading to a network that will %
be prone to feature collapse, unlike the DM-MLP.}
\label{fig:collapse}
\end{figure}

In order to illustrate the feature collapse problem, we consider a toy example on the two moon dataset. We train on it two MLPs with two hidden layers, each containing 17 neurons. 
One of the MLP additionally integrates our proposed DM layer and is denoted as DM-MLP, while the standard architecture is %
called MLP. The two networks reach the same classification performance, about 99\% of accuracy. We perform PCA on the pre-logit latent space of both networks after 
training and visualize PCA projections in Figure~\ref{fig:collapse}.
We can observe the feature collapse
as the MLP assigns strongly correlated feature representation to both classes which can lead to unreliable uncertainty prediction. 
However, our DM layer allows a better disentangling of the latent space. Note that, as the networks have the same performance, it is impossible to detect the feature collapse based on the test accuracy alone. 

\Gianninew{We note that our LDU layer is a Lipschitz function, hence: $\| \mbox{exp}(-\mbox{DM}_{p}(z_1)) -\mbox{exp}(-\mbox{DM}_{p}(z_2)) \| \leq k \|z_1-z_2 \|$ with $k \in \mathbb{R}^+$.
However, $h_{\vomega}$ is not necessarily a Lipschitz function, and we cannot thus guarantee that its features do not entangle ID and OOD data.
Yet, using a distance function in the DNN~\cite{mandelbaum2017distance,macedo2022distinction} can  allow it to learn to separate the two data distributions better as illustrated in
Figure~\ref{fig:collapse}.}

Most DUM methods 
aim for bi-Lipschitz DNNs with small Lipschitz constants. Yet, this is sub-optimal according to the concentration theory. Indeed, let $\vX$ be a set of random vectors of size $d$ i.i.d.  from a normal distribution $\mathcal{N}(0,\sigma^2I_d )$. $I_d$ is the identity matrix of size $d$.  Let $f: \mathbb{R}^d \rightarrow \mathbb{R}$ be a Lipschitz function with Lipschitz constant $K$. The concentration theory (\cite{boucheron2013concentration}, p. 125) stipulates that : 
$
\mathcal{P}(|f(\vX)-\mathbb{E}(f(\vX))| >t )\leq 2 \mbox{exp}(-\frac{t^2}{2K^2\sigma^2}) \mbox{ for all } t>0.
$
This means that the smaller $K$ is, the more the concentration of the data around their mean increases, leading to 
increased feature collapse. Hence, it is 
desirable to have a Lipschitz function that will bring similar data close, but it is at the same time essential to put dissimilar data apart.

\subsection{LDU and Epistemic/Aleatoric Uncertainty}\label{sec:LDU}

We are interested in capturing two types of uncertainty with our DNN: aleatoric and epistemic uncertainty~\cite{der2009aleatory,kendall2017uncertainties}. Aleatoric uncertainty is related to the inherent noise and ambiguity in data or in annotations and is often called irreducible uncertainty as it does not fade away with more data in the training set. Epistemic uncertainty is related to the model, the learning process and the amount of training data. Since it can be reduced with more training data, it is also called reducible uncertainty. Disentangling these two sources of uncertainty is generally non-trivial~\cite{mukhoti2021deterministicDDU} and ensemble methods are usually superior~\cite{fort2019deep,malinin2020ensemble}.

\begin{figure}[t]
\centering
\begin{tabular}{cc}
\tiny{\textbf{\begin{tabular}[c]{@{}c@{}}Confidence score\\after the first training\end{tabular}}} & \tiny{\textbf{\begin{tabular}[c]{@{}c@{}}Confidence score\\after the second training\end{tabular}}}\\
\includegraphics[width=0.20\linewidth]{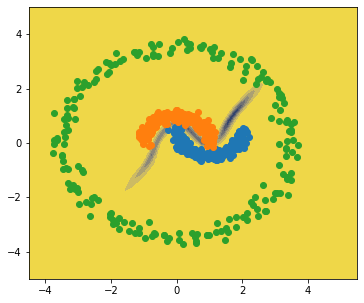} &
\includegraphics[width=0.20\linewidth]{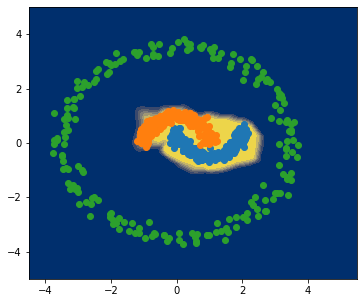}
\end{tabular}
\caption{Illustration of confidence score results on the two moons dataset after the first training \ab{(on original data)} on the left and with second training \ab{(on synthesized outliers)} on the right. 
Orange and blue data points are sampled from two classes in two moons, and the green points are OOD data points. Yellow area indicates high confidence, blue area indicates uncertainty. The left image shows that the uncertain area is between the two classes leading to a confidence score related to aleatoric uncertainty. In the right one, the uncertain area is around the dataset leading to a confidence score related to epistemic uncertainty.}
\label{fig:uncertainty}
\end{figure}
\noindent\comEmi{\textbf{Optional training with synthesized outliers. }} %
\comEmi{Due to limited training data and to the penalty enforced by $\mathcal{L}^{\mbox{Task}}$ being too small,  the loss term $\mathcal{L}^{\mbox{Unc}}$ may potentially force the DNN in some circumstances to overfit the aleatoric uncertainty.} 
\comEmi{ Although we did not encounter this behavior on the computer vision tasks given the dataset size, it might occur on more specific data, and among other potential solutions, we propose one relying on synthesized outliers that we illustrate on the two moons dataset as follows. }
\comEmi{More specifically, we propose to add noise to the data similarly to \cite{malinin2018predictive,du2022vos}, and to introduce an optional step for training $g_{\omega}^{\mbox{unc}} $ on these new samples.}
We consider a two-stage training scheme. In the first stage we train over data without noise and in the second we optimize only the parameters of $g_{\omega}^{\mbox{unc}} $ over the synthesized outliers. 
\comEmi{Note that this optional stage would require for vision tasks an adequate OOD synthesizer\ab{~\cite{besnier2021triggering,du2022vos}} which is beyond the scope of this paper, and that we applied it on the toy dataset.}
In Figure~\ref{fig:uncertainty} we assess the uncertainty estimation performance of this model on the two moons dataset. We can see that the confidence score relates to the aleatoric uncertainty after the first training stage. After the second one, it is linked to the epistemic uncertainty of the model.

Distinguishing between the two sources of uncertainty is 
\ab{essential}
for various practical applications, such as active learning, monitoring, OOD detection. In the following, we propose two strategies for computing each type of uncertainty.

\parag{Aleatoric uncertainty.} For estimating aleatoric uncertainty in classification, maximum class probability (MCP)~\cite{hendrycks2016baseline} is a common strategy. The intuition is that a lower MCP can mean a higher entropy, i.e., a potential confusion on the classifier regarding the most likely class of the image. We use this criterion for the aleatoric uncertainty for classification and for semantic segmentation, while for the regression task we use $g_{\omega}^{\mbox{unc}} $ as confidence score.

\parag{Epistemic uncertainty.} To estimate epistemic uncertainty, we analyzed the latent representations of the DM layers followed by an exponential activation and found that the 
maximum value can model well uncertainty. The position of the feature w.r.t. the learned prototypes caries information about the proximity of the current sample with the in-distribution features. Yet, we propose to use the output of $g_{\omega}^{\mbox{unc}} $ as confidence score since we train this criterion for this purpose.

\section{Experiments}\label{sec:Experiments}

One major interest of our technique is that it may be seamlessly applied to any computer vision task, be it classification or regression. Thus, we propose to evaluate the quality of uncertainty quantification of different techniques on three major tasks, namely image classification, semantic segmentation and monocular depth estimation. 
For 
\ab{all the} three tasks, we compare our technique against MC Dropout \cite{gal2016dropout} and Deep Ensembles \cite{lakshminarayanan2017simple}. For image classification, we also compare our technique to relevant DUM techniques for image classification, namely  DDU \cite{mukhoti2021deterministicDDU}, DUQ \cite{van2020uncertainty}, DUE \cite{van2021improvingDUE}, MIR \cite{postels2020hiddenMIR} and SNGP \cite{liu2020simpleSNGP}.

We evaluate the predictive performance in terms of accuracy for image classification, mIoU~\cite{everingham2015pascal} for semantic segmentation, and the %
metrics first introduced in~\cite{NIPS2014_7bccfde7} and used in many subsequent works for monocular depth estimation. 
For image classification and semantic segmentation, we also evaluate the quality of the confidence scores provided by the DNNs via the following metrics: Expected Calibration Error (ECE)~\cite{guo2017calibration}, AUROC~\cite{hendrycks2016baseline,hendrycks2019anomalyseg} and AUPR~\cite{hendrycks2016baseline,hendrycks2019anomalyseg}. Note that ECE we use is the confidence ECE defined in~\cite{guo2017calibration}.
We use the Area Under the Sparsification Error: AUSE-RMSE and AUSE-Absrel similarly to~\cite{poggi2020uncertainty,gustafsson2020evaluating,yu2021slurp} to better evaluate the uncertainty quantification on monocular depth estimation. 

\Xuanlong{We run all methods ourselves in similar settings using publicly available codes \ab{and hyper-parameters} for related methods. In the following tables, Top-2 results are highlighted in color.}

\subsection{Classification experiments}

 To evaluate uncertainty quantification for image classification, we adopt a standard approach based on training on CIFAR-10~\cite{krizhevsky2009learning}, 
\ab{and using} SVHN~\cite{Netzer2011} as OOD data~\cite{postels2021practicality,franchi2020tradi,liu2020simpleSNGP}.
We use  ResNet18 \cite{he2016deep} as architecture for all methods.
Note that for all DNNs, even for DE,  we average results over three random seeds for statistical relevance. We follow 
\ab{the corresponding} protocol for all DUM techniques (except LDU). For Deep Ensembles, MCP, and LDU, we use the same protocol. 
\ab{Please refer to the appendix for implementation details of LDU.}
The performances of the different algorithms are shown in Table \ref{table:tab1}. We can see that LDU has state-of-the-art performances on CIFAR-10. 
We note that LDU's OOD detection performance improves with the number of prototypes.
This can be linked with the fact that the more prototypes we have, the better we can model complex distributions. \Xuanlong{The ablation studies on sensitivity of the choice of $\lambda$ and the impact of different losses are provided in the 
\ab{appendix}.}

\begin{table}[!t]
\setlength{\abovecaptionskip}{0.cm}
\renewcommand{\figurename}{Table}
\begin{center}
\scalebox{0.55}
{
\begin{tabular}{l|cccc}%
\toprule
 &   \multicolumn{4}{c}{CIFAR-10} \\
Method                  & Acc $\uparrow$   & AUC $\uparrow$  & AUPR $\uparrow$     & ECE $\downarrow$ \\
\midrule
Baseline (MCP) ~\cite{hendrycks2016baseline}             & 88.02 & 0.8032 & 0.8713 & 0.5126   \\
\midrule
MCP lipz. ~\cite{behrmann2019invertible}               & 88.50 & 0.8403 & 0.9058 & \first \textbf{0.3820}   \\
\midrule
Deep Ensembles~\cite{lakshminarayanan2017simple}               & \first \textbf{89.96} & 0.8513 &	0.9087&	0.4249  \\
\midrule
SNGP \cite{liu2020simpleSNGP}         & 88.45 &	0.8447 &	0.9139 &	0.4254  \\
\midrule
DUQ \cite{van2020uncertainty}          & \second 89.9 &	0.8446 &\second	0.9144 &	0.5695  \\
\midrule
DUE \cite{van2021improvingDUE}          & 87.54 &	0.8434 &	0.9082 &	0.4313  \\
\midrule
DDU  \cite{mukhoti2021deterministicDDU}       & 87.87 &	0.8199 &	0.8754  & \first \textbf{0.3820}  \\
\midrule
MIR  \cite{postels2020hiddenMIR}          &  87.95 &	0.7574 &	0.8556 &	\second 0.4004   \\
\midrule
\midrule
LDU $\# \vp  = 128$            & 87.95	& \first \textbf{0.8721} & \first \textbf{0.9147} & 0.4933   \\
LDU $\# \vp  = 64$            & 88.06 & \second 0.8625 & 0.9070 & 0.5010    \\
LDU $\# \vp  = 32$            & 87.83 & 0.8129 & 0.8900 & 0.5264    \\
LDU $\# \vp  = 16$            & 88.33 & 0.8479 & 0.9094 & 0.4975   \\
\bottomrule
\end{tabular}
} %
\end{center}
\caption{Comparative results for \textbf{image classification tasks}. We evaluate on CIFAR-10 for the tasks: in-domain classification, out-of-distribution detection with SVHN. 
Results are averaged over three seeds.}
\vspace{-1.5em}
\label{table:tab1}
\end{table}

\begin{figure}[t]
\centering
\begin{tabular}{ccccc}
Image& groundtruth & prediction & MCP & $g_{\omega}^{\mbox{unc}}$'s prediction\\
\includegraphics[width=0.20\linewidth]{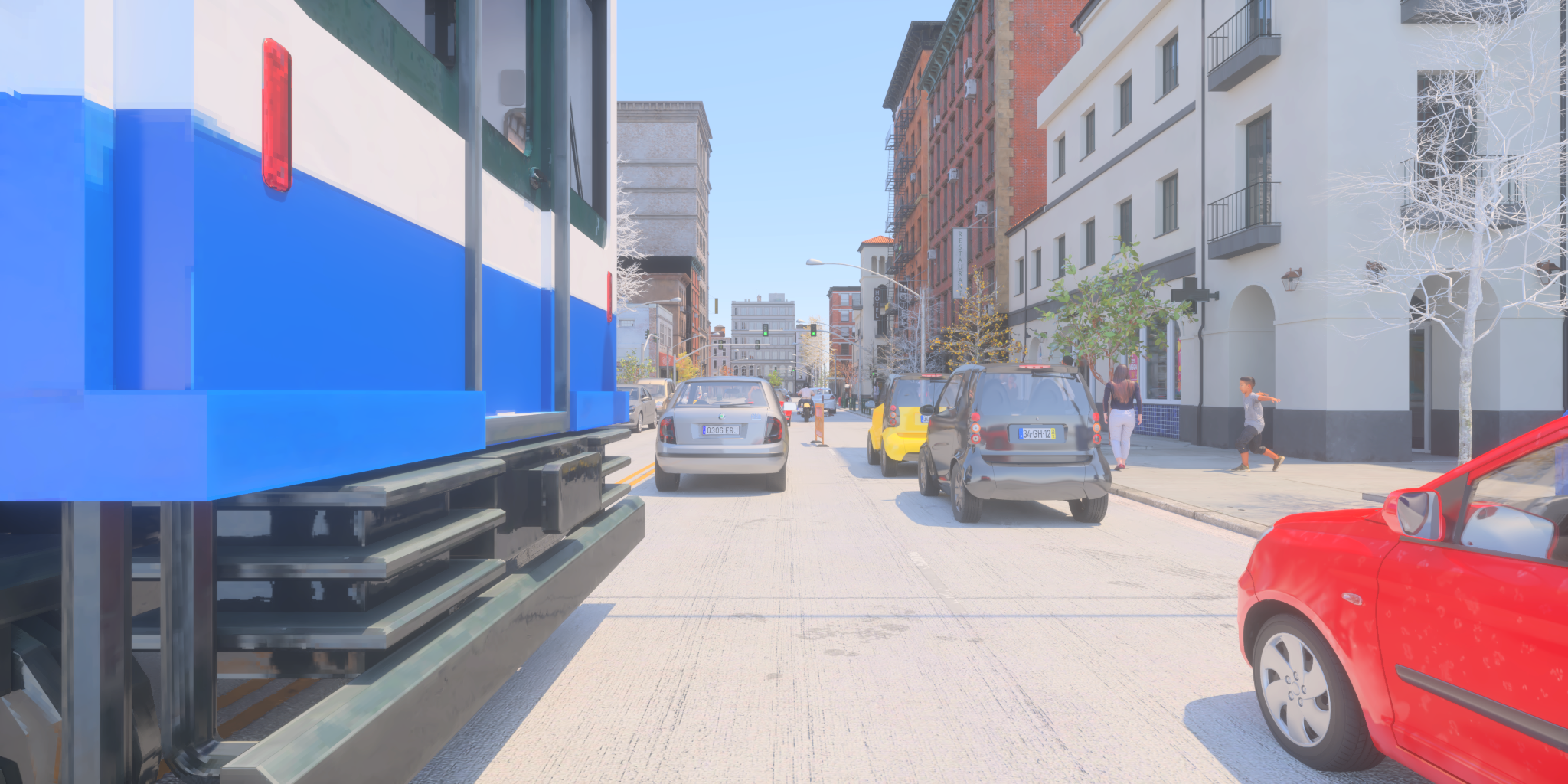}&
\includegraphics[width=0.20\linewidth]{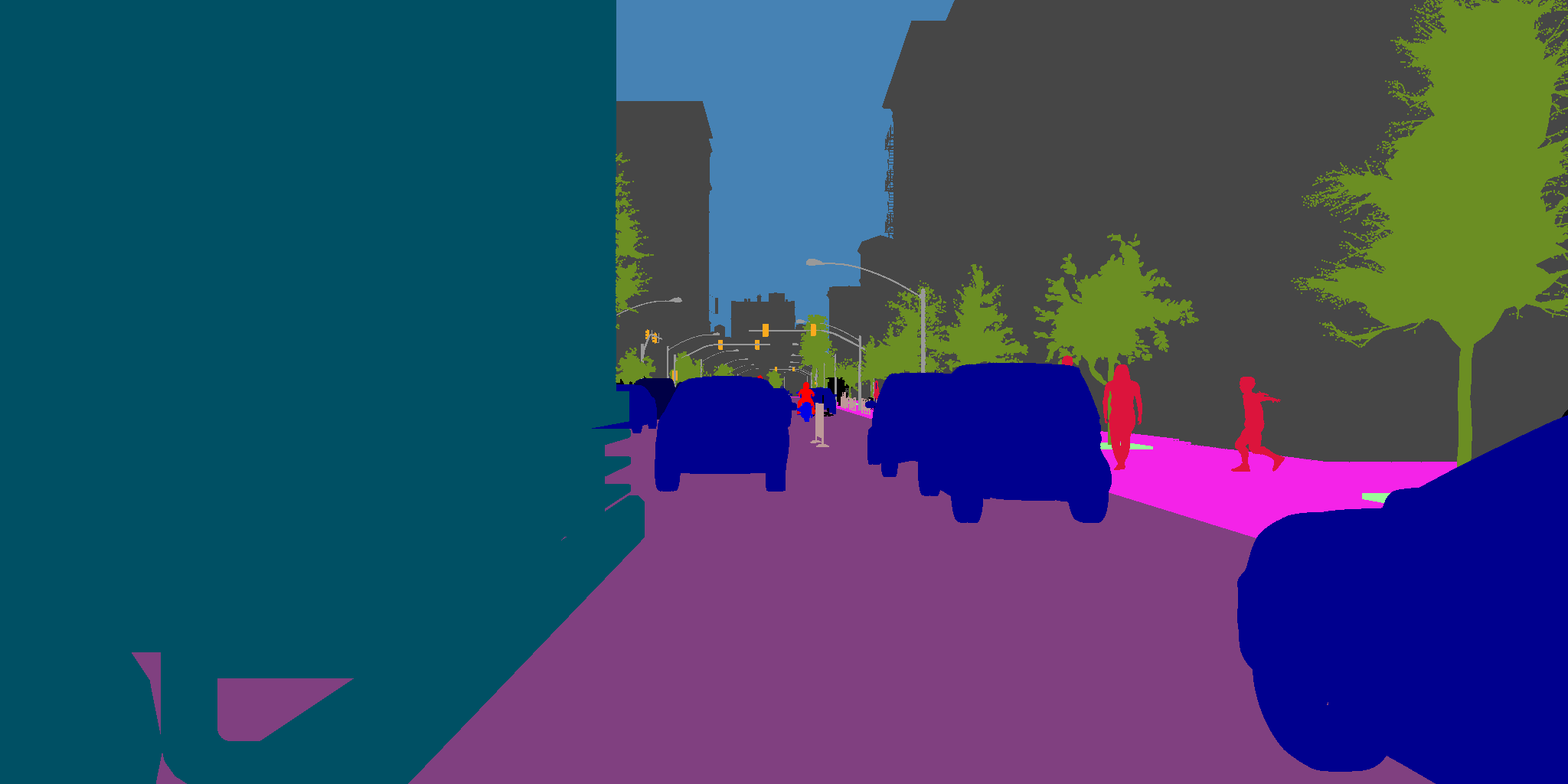}&
\includegraphics[width=0.20\linewidth]{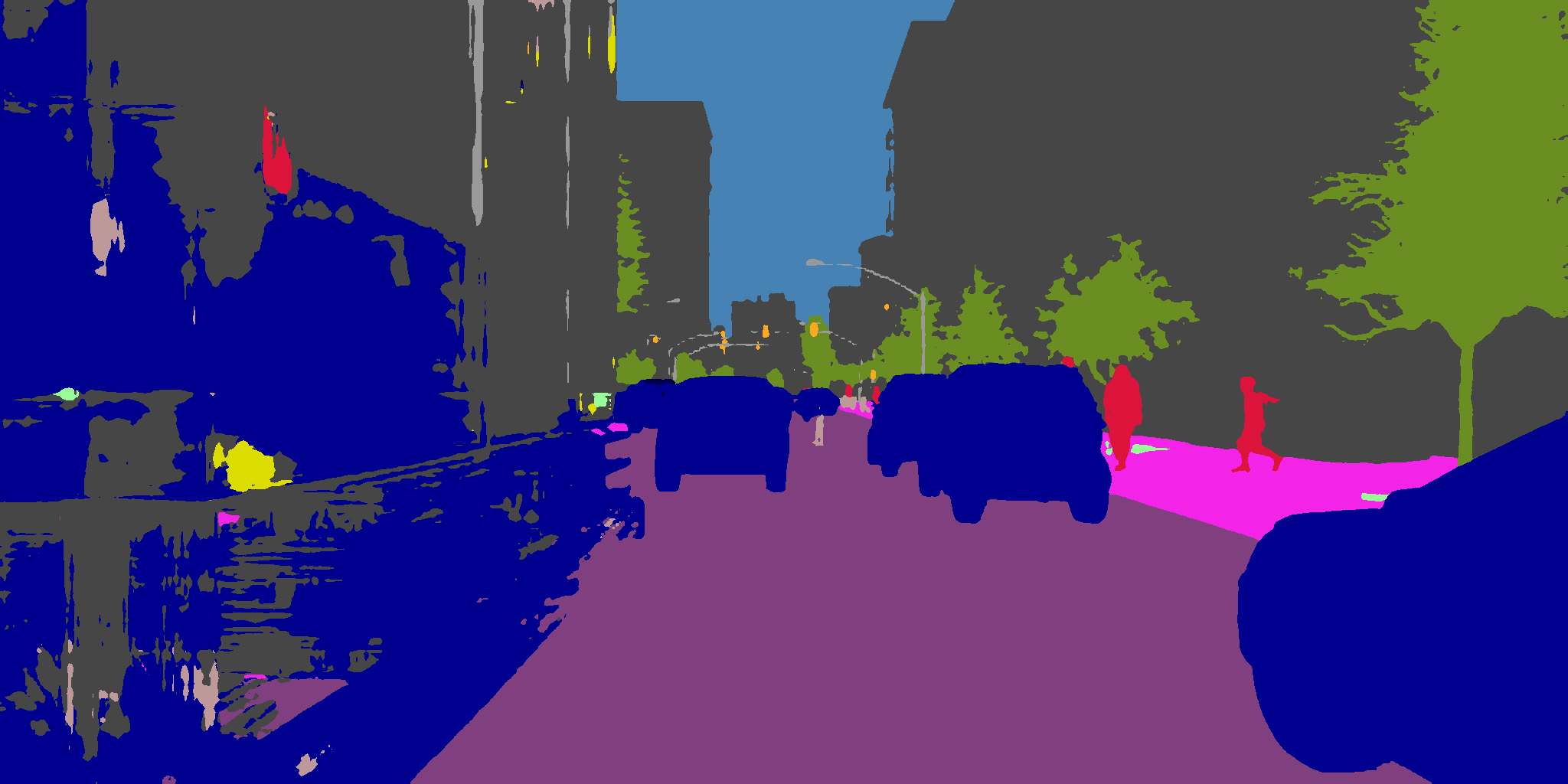}&
\includegraphics[width=0.20\linewidth]{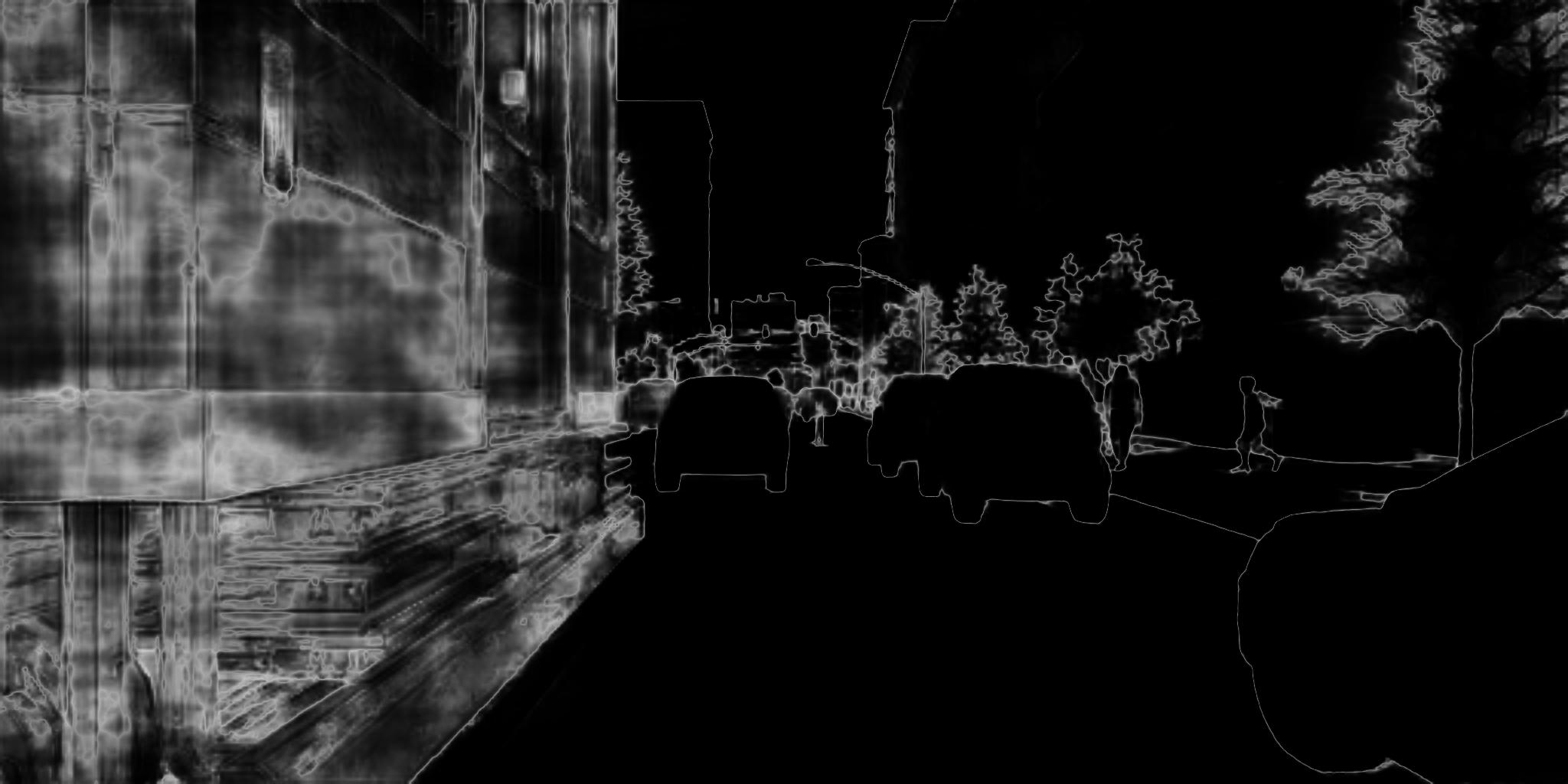}&
\includegraphics[width=0.20\linewidth]{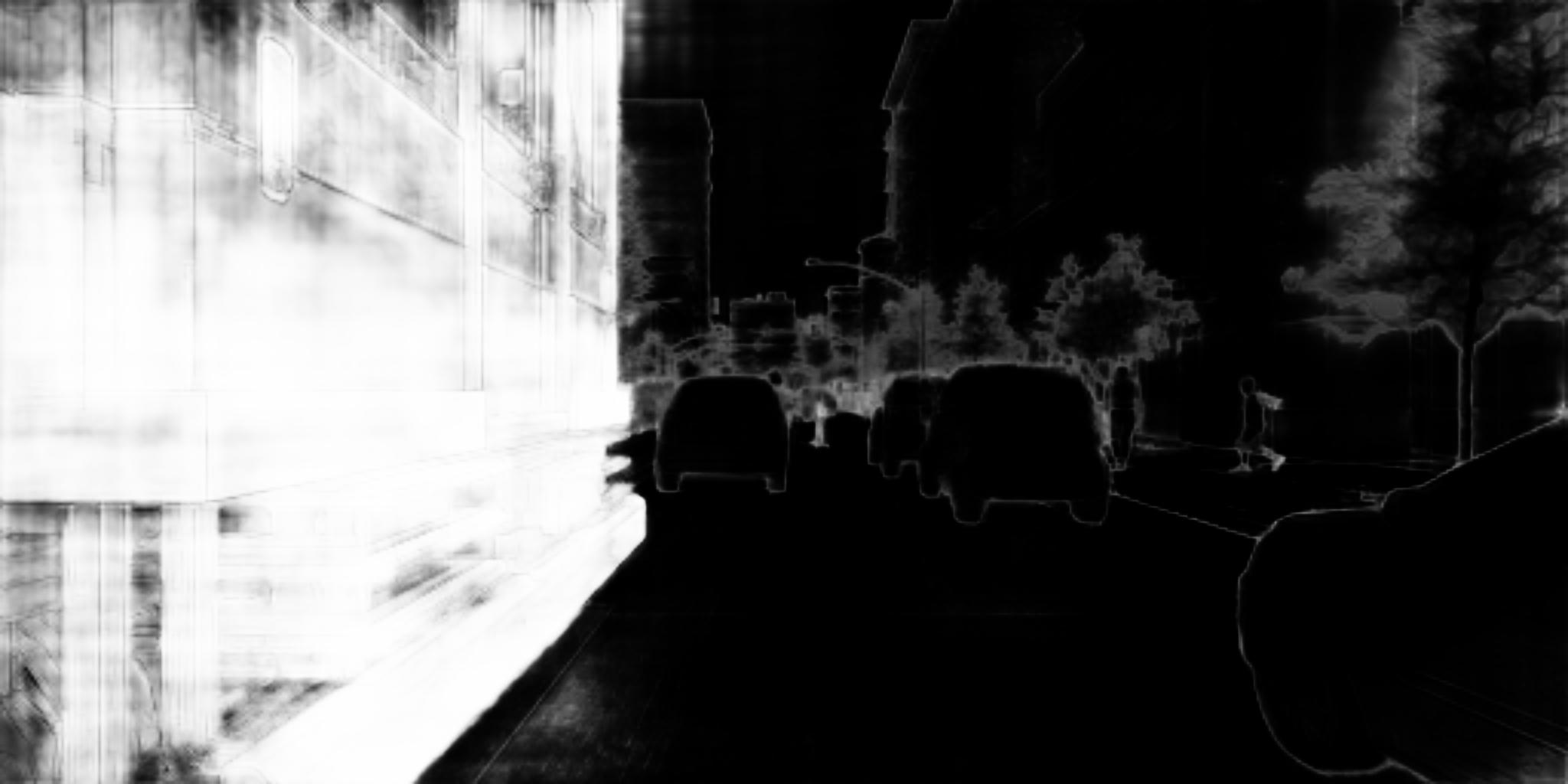}\\
\includegraphics[width=0.20\linewidth]{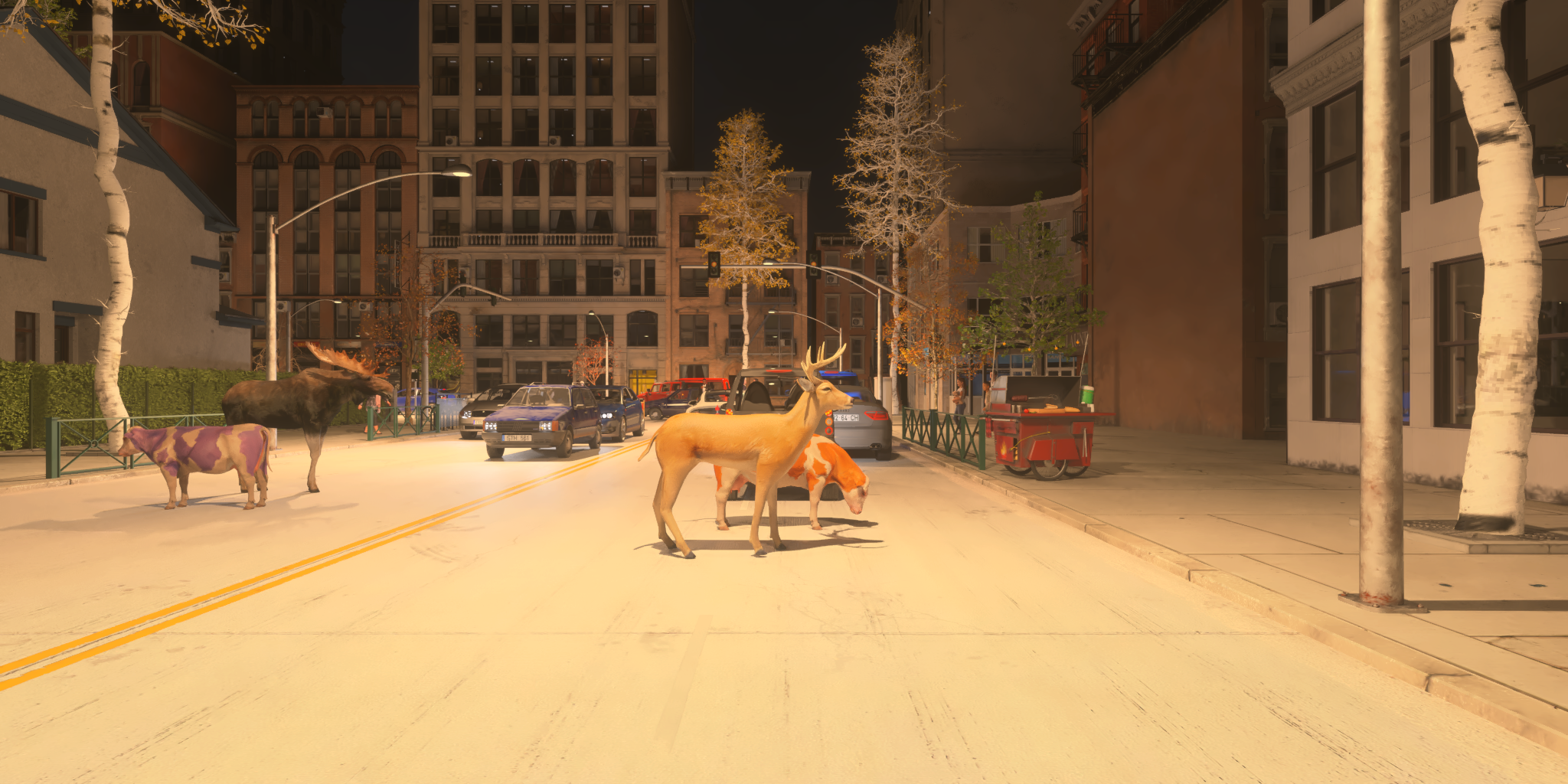}&
\includegraphics[width=0.20\linewidth]{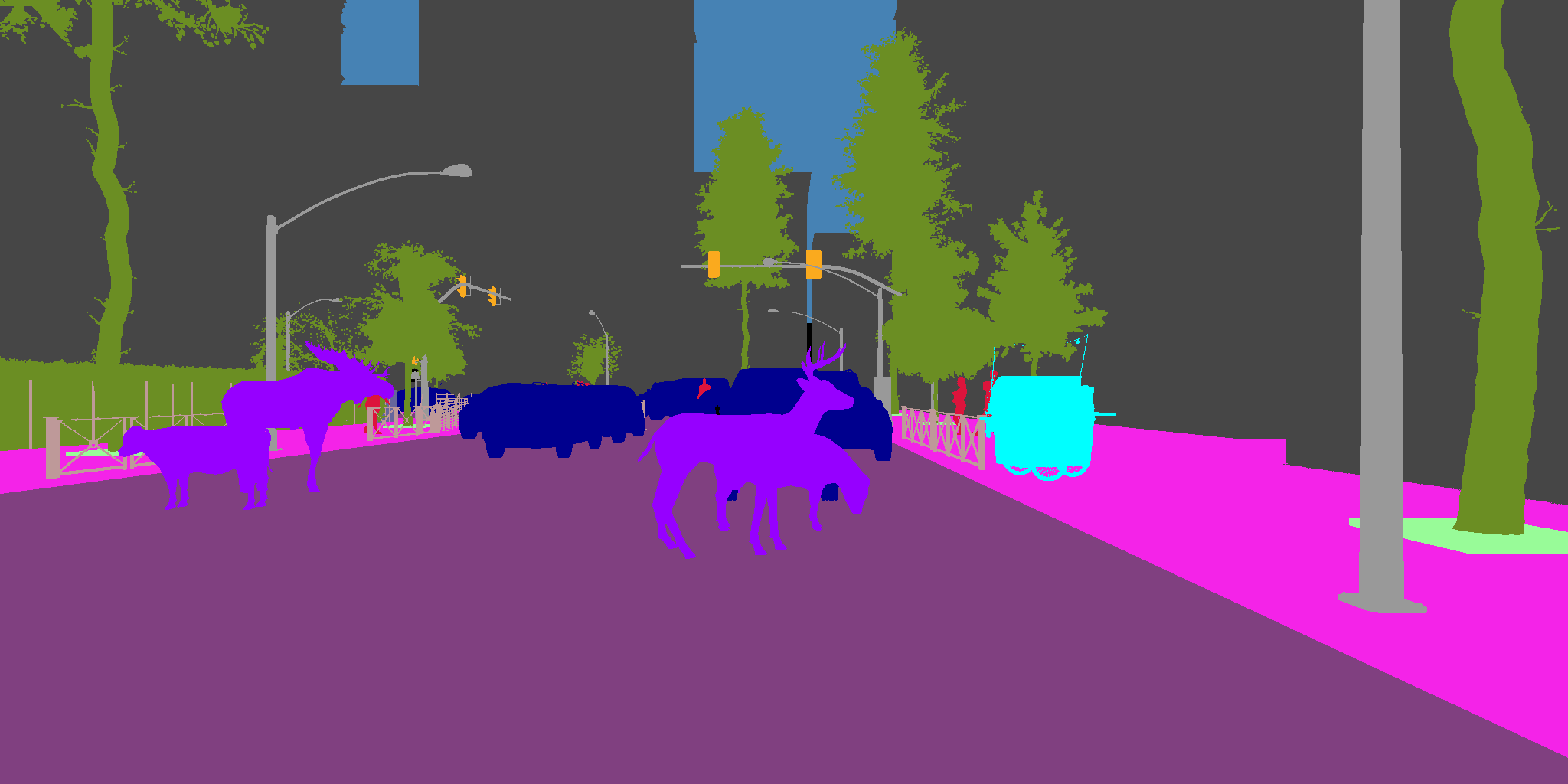}&
\includegraphics[width=0.20\linewidth]{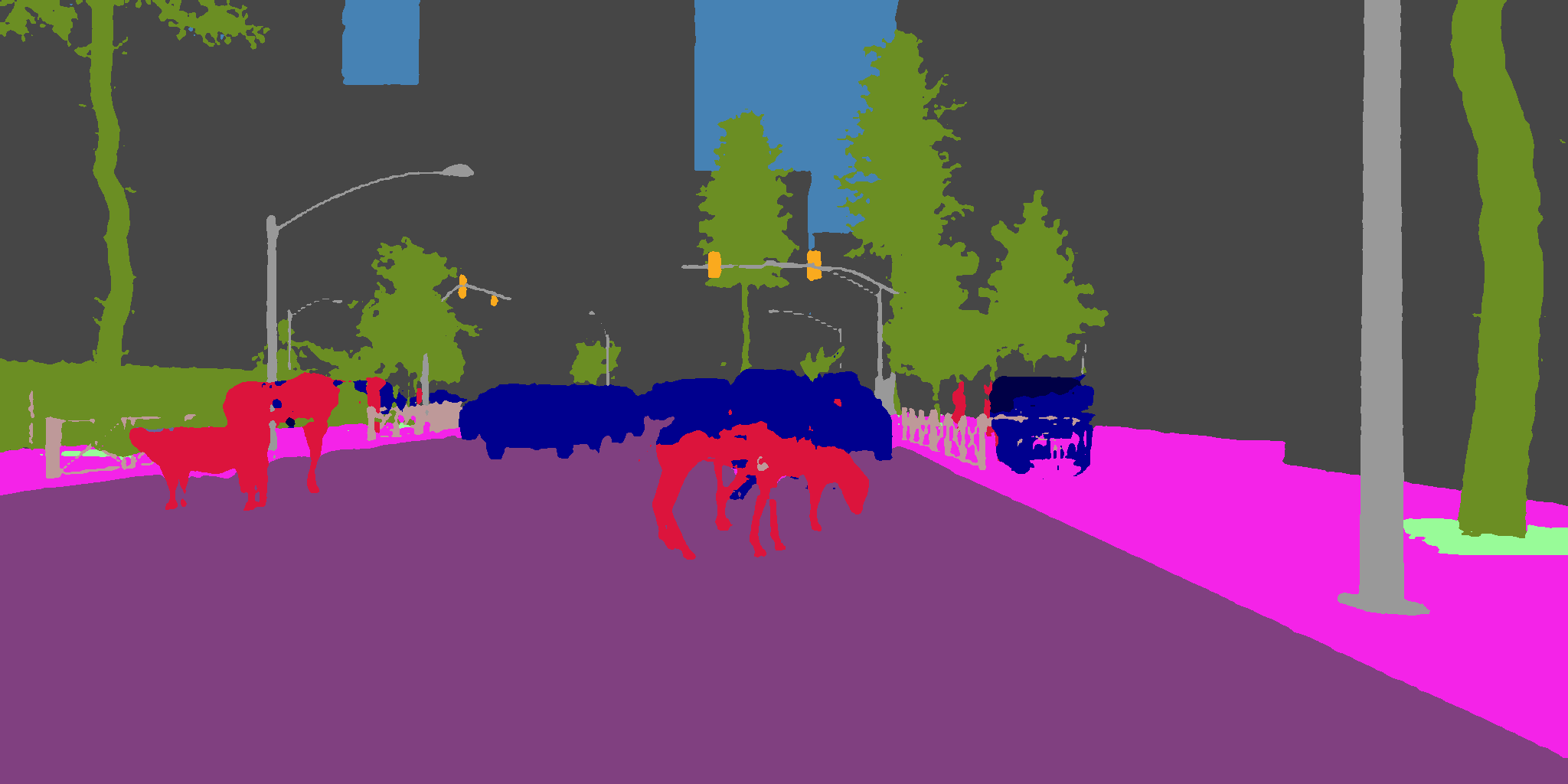}&
\includegraphics[width=0.20\linewidth]{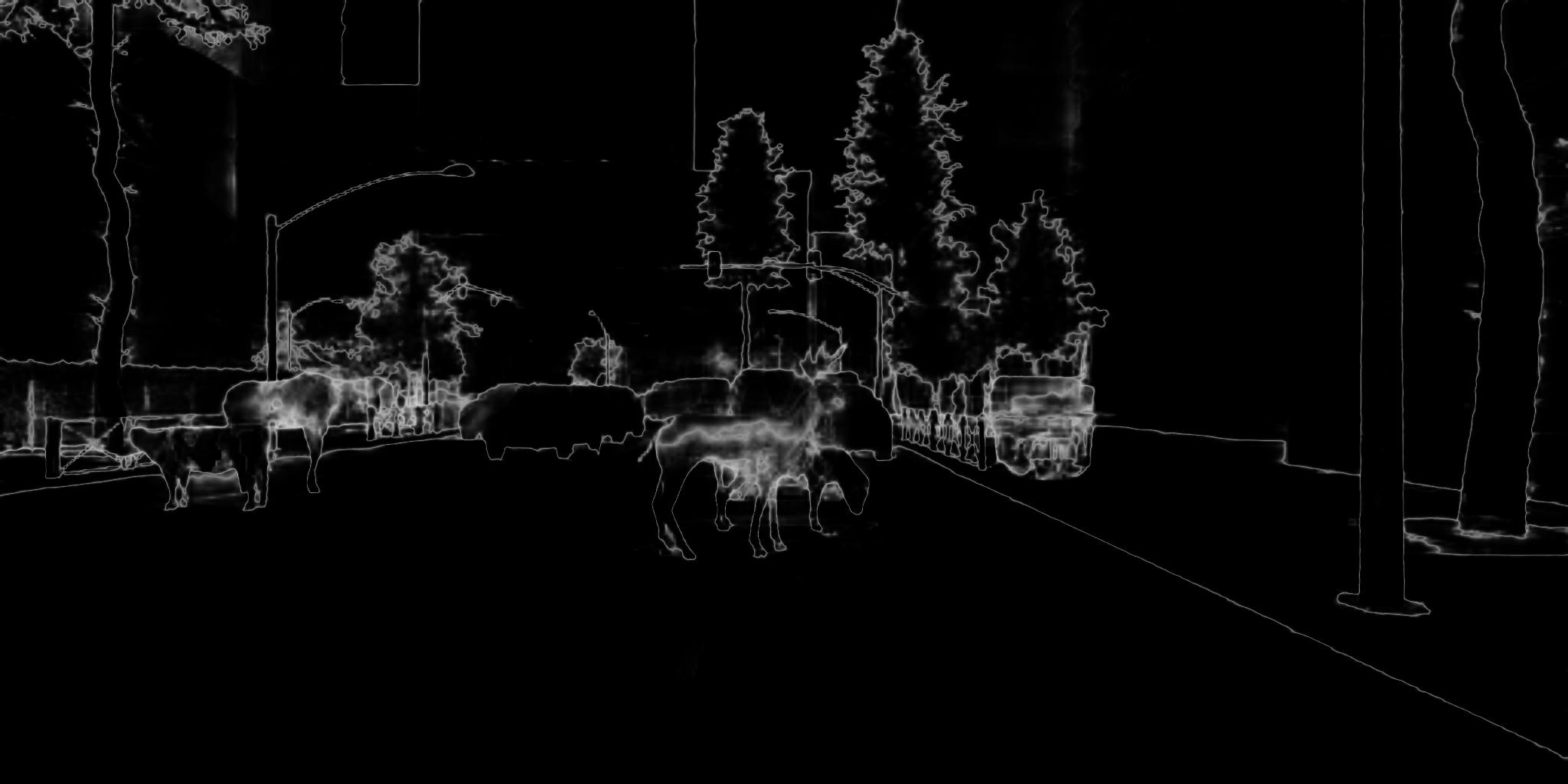}&
\includegraphics[width=0.20\linewidth]{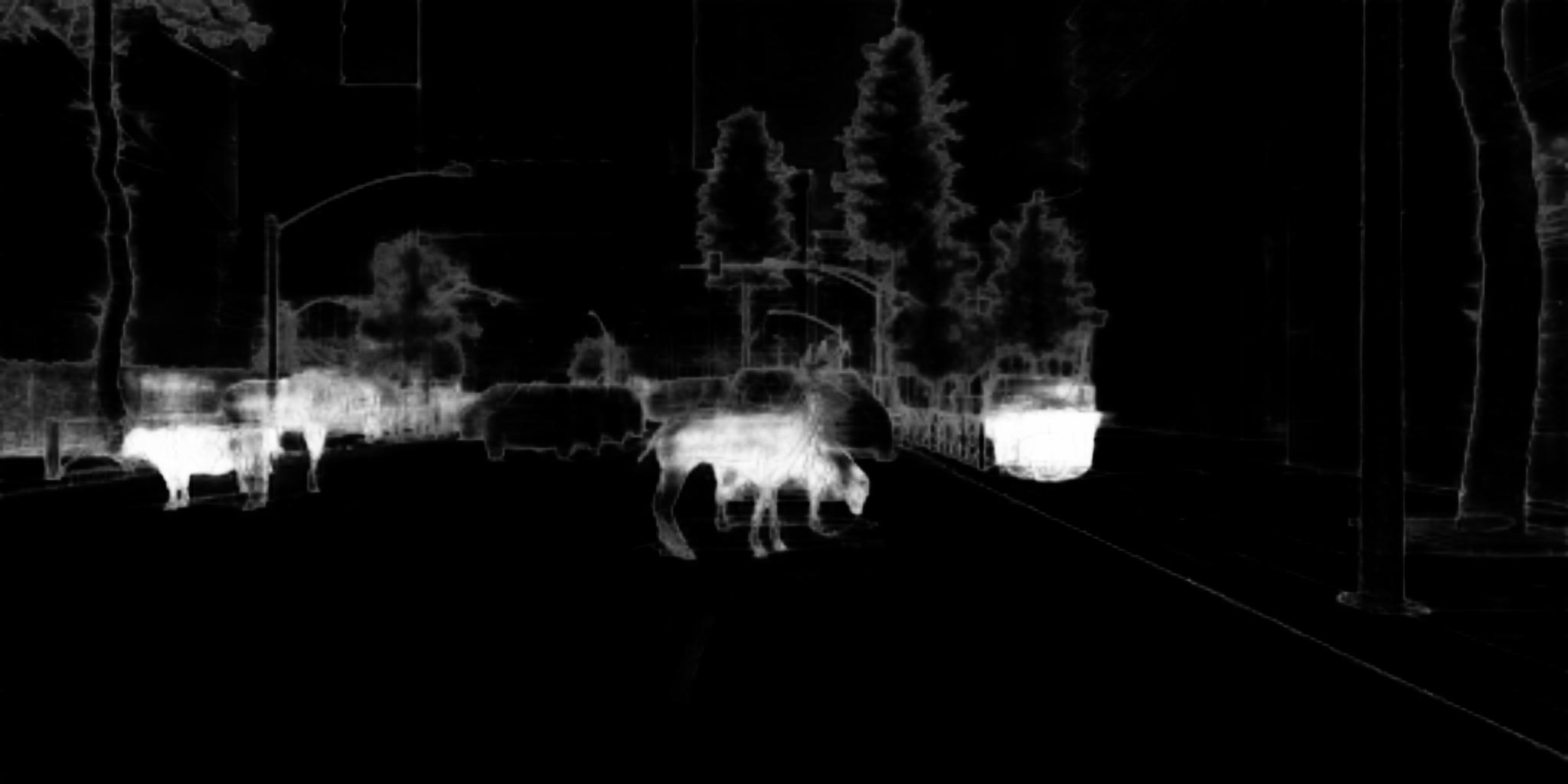}\\
\end{tabular}
\caption{Illustration of the different confidence scores on one image of MUAD. Note that the class \texttt{train}, \texttt{bicycle}, \texttt{Stand food} and the \texttt{animals} are OOD.}
\vspace{-1em}
\label{fig:conf}
\end{figure}
\subsection{Semantic segmentation experiments}\label{sec:segment_exp}

\begin{table*}[t]
\setlength{\abovecaptionskip}{0.cm}
\begin{center}
 \scalebox{0.45}
 {
\begin{tabular}{lcc|ccccc|ccccc|ccccc} 
\toprule
\multicolumn{1}{c}{\multirow{2}{*}{Evaluation data}} & \multicolumn{2}{c}{\textbf{normal set}} & \multicolumn{5}{c}{\textbf{OOD set}}  & \multicolumn{5}{c}{\textbf{low adv. set}}  & \multicolumn{5}{c}{\textbf{high adv. set}}  \\
    &  mIoU $\uparrow$ &  ECE $\downarrow$   &  mIoU $\uparrow$  &  ECE $\downarrow$ &  AUROC $\uparrow$ &  AUPR $\uparrow$ &  FPR $\downarrow$  &  mIoU $\uparrow$ &  ECE $\downarrow$  &  AUROC $\uparrow$ &  AUPR $\uparrow$ &  FPR $\downarrow$   & mIoU $\uparrow$    & ECE $\downarrow$ &   AUROC $\uparrow$ &  AUPR $\uparrow$ &  FPR $\downarrow$  \\
\midrule
Baseline (MCP)~\cite{hendrycks2016baseline} &  68.90\% & 0.0138 & \second 57.32\% & 0.0607 & 0.8624 & 0.2604 & 0.3943 & 31.84\% & 0.3078 & 0.6349 & 0.1185 & 0.6746 & 18.94\% & 0.4356 & 0.6023 & 0.1073 & 0.7547\\
\midrule
 Baseline (MCP) lipz.~\cite{behrmann2019invertible} &  53.96\%  & 0.01398 & 45.97\%  & 0.0601 & 0.8419 & 0.2035 & 0.3940 & 16.79\%  & 0.3336 & 0.6303 & 0.1051 & 0.7262 & 7.8\%  & 0.4244 & 0.5542 & 0.0901 & 0.8243\\
 \midrule
MIR \cite{postels2020hiddenMIR} & 
53.96\%  & 0.01398 & 45.97\%  & 0.0601 & 0.6223 & 0.1469 & 0.8406 & 16.79\%  & 0.3336 & 0.5143 & 0.1035 & 0.8708 & 7.8\%  & 0.4244 & 0.4470 & 0.0885 & 0.9093\\
  \midrule
MC-Dropout \cite{gal2016dropout} &  65.33\% & 0.0173 & 55.62\% & 0.0645 & 0.8439 & 0.2225 & 0.4575 & 33.38\% & \first \textbf{0.1329} & \second 0.7506 &  \second 0.1545 &  \second 0.5807 &  \second 20.77\% &  \first \textbf{0.3809} &  \second 0.6864 &  \second 0.1185 &  \second 0.6751\\
 \midrule
Deep Ensembles ~\cite{lakshminarayanan2017simple} &  \first \textbf{69.80\%} & \first \textbf{0.01296} & \first \textbf{58.29\%} & \first \textbf{0.0588} &  \second 0.871 & \second 0.2802 & \second 0.3760 & \second 34.91\% & \second 0.2447 & 0.6543 & 0.1212 & 0.6425 & 20.19\% & \second 0.4227 & 0.6101 & 0.1162 & 0.7212\\
 \midrule
LDU (ours) &  \second 69.32\% & \second 0.01356 & \first  \textbf{58.29\%} & \second 0.0594 & \first \textbf{0.8816} & \first \textbf{0.4418} & \first \textbf{0.3548} & \first\textbf{ 36.12\%} & 0.2674 & \first \textbf{0.7779} & \first \textbf{0.2898} & \first \textbf{0.5381} & \first \textbf{ 21.15\%} & 0.4231 & \first \textbf{0.7107} & \first \textbf{0.2186} &\first \textbf{0.6412}\\
\bottomrule
\end{tabular}
}
\end{center}
\vspace{-2mm}
\caption{{Comparative results for \textbf{semantic segmentation on MUAD}}.
}
\label{table:tnb2}
\vspace{-2mm}
\end{table*}

\begin{table*}[t]
\setlength{\abovecaptionskip}{0.cm}
\begin{center}
 \scalebox{0.5}
 {
\begin{tabular}{lcc|cc|cc|cc|cc|cc} 
\toprule
\multicolumn{1}{c}{\multirow{2}{*}{Evaluation data}} & \multicolumn{2}{c}{\textbf{Cityscapes}} & \multicolumn{2}{c}{\textbf{Cityscapes-C lvl 1}}  & \multicolumn{2}{c}{\textbf{Cityscapes-C lvl 2}}  & \multicolumn{2}{c}{\textbf{Cityscapes-C lvl 3}}& \multicolumn{2}{c}{\textbf{Cityscapes-C lvl 4}}  & \multicolumn{2}{c}{\textbf{Cityscapes-C lvl 5}}  \\
    &  mIoU $\uparrow$ &  ECE $\downarrow$   &  mIoU $\uparrow$  &  ECE $\downarrow$   &  mIoU $\uparrow$ &  ECE $\downarrow$    & mIoU $\uparrow$    & ECE $\downarrow$  &  mIoU $\uparrow$ &  ECE $\downarrow$    & mIoU $\uparrow$    & ECE $\downarrow$\\
\midrule
Baseline (MCP)~\cite{hendrycks2016baseline} &  \second 76.84\% & 0.1180 & 51.59\% & 0.1793 & 41.45\% & 0.2291 &35.67\% & 0.2136 & 30.12\% & 0.1970 & 24.84\% & 0.2131\\
\midrule
 Baseline (MCP) lipz.~\cite{behrmann2019invertible}& 58.38\% & \second 0.1037 & 44.70\% & \first \textbf{0.1211} & 38.04\% & \second 0.1475 & 32.70\% & 0.1802 & 25.35\% & 0.2047 & 18.36\% & 0.2948\\
 \midrule
MC-Dropout \cite{gal2016dropout} & 71.88\% & 0.1157 & \second 53.61\% & 0.1501 & 42.02\% & 0.2531 & 35.91\% & \second 0.1718 & 29.52\% & 0.1947 & 25.61\% & 0.2184\\
 \midrule
Deep Ensembles ~\cite{lakshminarayanan2017simple} &  \first\textbf{77.23\%} & 0.1139 & \first \textbf{54.98\%} & 0.1422 & \first \textbf{44.63\%} &  0.1902 & \first \textbf{38.00\%} & 0.1851 & \second 32.14\% & \second0.1602 & \first \textbf{28.74\%} & \second 0.1729\\
 \midrule
LDU (ours)  &  76.62\% & \first\textbf{0.0893} & 52.00\% & \second 0.1371 & \second 43.02\% & \first \textbf{0.1314} & \second 37.17\% & \first \textbf{0.1702} & \first \textbf{32.27\%} & \first \textbf{0.1314} & \second 27.30\% & \first \textbf{0.1712}\\
\bottomrule
\end{tabular}
}
\end{center}
\vspace{-2mm}
\caption{{Comparative results for \textbf{semantic segmentation on Cityscapes and Cityscapes-C}}.
}
\label{table:tnb2_supp}
\vspace{-2mm}
\end{table*}

\begin{table*}[th!]
\begin{center}
\resizebox{0.45\linewidth}{!}
{
\begin{tabular}{l r r r r }
\toprule
 OOD technique  & mIoU $\uparrow$  & AUC $\uparrow$  & AUPR $\uparrow$ & FPR-95\%-TPR $\downarrow$\\ \midrule
Baseline (MCP) ~\cite{hendrycks2016baseline}& 52.8 & 86.0 & 5.4  & 27.7         \\  
 MC-Dropout  \cite{gal2016dropout}   & 49.5  & 85.2 & 5.0  & 29.3         \\  
Deep Ensembles~\cite{lakshminarayanan2017simple} & \first \textbf{57.6} & \second 87.0 & \first \textbf{ 6.0 } & \first \textbf{25.0}         \\  
 TRADI \cite{franchi2020tradi}      & 52.1   & 86.1 & 5.6  & 26.9         \\  
ConfidNET \cite{corbiere2019addressing}    & 52.8  & 85.4 & 5.1  & 29.1         \\ 
LDU  (ours) & \second 55.1  & \first \textbf{87.1} & \second 5.8  & \second 26.2         \\ 
\bottomrule
\end{tabular}
} %
\end{center}
\caption{{Comparative results obtained on the \textbf{OOD detection task on BDD Anomaly}}~\cite{hendrycks2019anomalyseg} with PSPNet (ResNet50).
}\label{table:OODresults2}
\vspace{-2em}
\end{table*}

\begin{table*}[t]
\begin{center}
\scalebox{0.45}
{
\begin{tabular}{llcccccccclcc} 
\toprule
\multirow{2}{*}{Method} && \multicolumn{8}{c}{\textbf{Depth performance}} &  & \multicolumn{2}{c}{\textbf{Uncertainty performance}} \\ 
\cmidrule{3-10}\cmidrule{12-13}
 && d1$\uparrow$ & d2$\uparrow$ & d3$\uparrow$ & Abs Rel$\downarrow$ & Sq Rel$\downarrow$ & RMSE$\downarrow$ & RMSE log$\downarrow$ & log10$\downarrow$ &  & AUSE RMSE$\downarrow$ & AUSE Absrel$\downarrow$ \\ 
\toprule
Baseline && \second 0.955 & \first \textbf{0.993} & \second 0.998 & \first \textbf{0.060} & 0.249 & 2.798 & \second 0.096 & \second 0.027 &  & - & - \\ 
\midrule
Deep Ensembles~\cite{lakshminarayanan2017simple}  && \first \textbf{0.956} & \first \textbf{0.993} & \first \textbf{0.999} & \first \textbf{0.060} & \first \textbf{0.236} & \first \textbf{2.700} & \first \textbf{0.094} & \first \textbf{0.026} &  & \first \textbf{0.08} & \first \textbf{0.21} \\ 
\midrule
MC-Dropout~\cite{gal2016dropout} && 0.945 & \second 0.992 & \second 0.998 & 0.072 & 0.287 & 2.902 & 0.107 & 0.031 &  & 0.46 & 0.50 \\ 
\midrule
Single-PU~\cite{kendall2017uncertainties} && 0.949 & 0.991 & \second 0.998 & 0.064 & 0.263 & 2.796 & 0.101 & 0.029 &  & \first \textbf{0.08} & \first \textbf{0.21} \\ 
\midrule
Infer-noise~\cite{mi2019training} && \second 0.955 & \first \textbf{0.993} & \second 0.998 & \first \textbf{0.060} & 0.249 & 2.798 & \second 0.096 & \second 0.027 &  & 0.33 & 0.48 \\ 
\midrule
LDU $\# \vp = 5, \hspace{0.2em}\lambda = 1.0$ && 0.954 & \first \textbf{0.993} & \second 0.998 & 0.063 & 0.253 & 2.768 & 0.098 & \second 0.027 &  & \first \textbf{0.08} & \first \textbf{0.21} \\ 
\midrule
LDU $\# \vp = 15, \hspace{0.2em}\lambda = 0.1$ && 0.954 & \first \textbf{0.993} & \second 0.998 & 0.062 & 0.249 & 2.769 & 0.098 & \second 0.027 &  & 0.10 & 0.28 \\ 
\midrule
LDU $\# \vp = 30, \hspace{0.2em}\lambda = 0.1$  && \second 0.955 & \second 0.992 & \second 0.998 & \second 0.061 & \second 0.248 & \second 2.757 & 0.097 & \second 0.027 &  & \second 0.09 & \second 0.26 \\
\bottomrule
\end{tabular}
}
\vspace{1em}
\caption{{Comparative results for \textbf{monocular depth estimation on KITTI eigen-split validation set}.} 
}
\vspace{-3em}
\label{table:depth_kitti}
\end{center}
\end{table*}

Our semantic segmentation study consists of three experiments. The first one is on a new synthetic dataset: MUAD~\cite{Franchi2022MUAD}. It comprises a training set and a test set without OOD classes and adverse weather conditions. We denote this set \textbf{normal set}. MUAD contains three more test sets that we denote \textbf{OOD set}, \textbf{low adv. set} and \textbf{high adv. set} which contain respectively 
images with OOD pixels but without adverse weather conditions, images with OOD pixels and weak adverse weather conditions, and for the last set,  images with OOD pixels and strong adverse weather conditions.
The second experiment evaluates the segmentation precision and uncertainty quality on the Cityscapes~\cite{cordts2016cityscapes} and the Cityscapes-C~\cite{kamann2021benchmarking,rebut2021styleless,franchi2021robust} datasets \ab{to assess performance under distribution shift}.
\ab{Finally} we 
\ab{analyze} OOD detection \ab{performance} on BDD Anomaly dataset~\cite{hendrycks2019anomalyseg} \ab{whose test set contains objects unseen during training.} 
\ab{We detail the experimental protocol of all datasets in the appendix.}

We train a DeepLabV3+~\cite{chen2018encoder} network with ResNet50 encoder~\cite{he2016deep} on MUAD. 
Table~\ref{table:tnb2}  lists the results from different uncertainty techniques. 
\ab{For} this task, we found that enforcing the Lipschitz constraint (see  Baseline (MCP) lipz.) %
has a significant impact.
Figure~\ref{fig:conf} shows \ab{a qualitative example of typical uncertainty maps computed on MUAD images.}

Similarly to \ab{\cite{franchi2021robust,kamann2021benchmarking}}  
we assess predictive uncertainty and robustness under distribution shift using Cityscapes-C, a corrupted version of Cityscapes images with perturbations of varying intensity. We generate Cityscapes-C ourselves from the original Cityscapes images using the code of Hendrycks et al.~\cite{hendrycks2019benchmarking} to apply the different corruptions on the images.
Following~\cite{hendrycks2019benchmarking}, we apply the following perturbations: Gaussian noise, shot noise, impulse noise, defocus blur, frosted, glass blur, motion blur, zoom blur, snow, frost, fog, brightness, contrast, elastic, pixelate, JPEG. 
Each perturbation is scaled with five levels of strength.  We train a DeepLabV3+~\cite{chen2018encoder} 
with ResNet50 encoder~\cite{he2016deep} on Cityscapes.
Results in Table~\ref{table:tnb2_supp} show that LDU is closely trailing in accuracy (mIoU score) the much more costly Deep Ensembles~\cite{lakshminarayanan2017simple}, while making better calibrated predictions (ECE score).

In order to assess the epistemic uncertainty quantification on real data we used PSPNet~\cite{zhao2017pyramid} \ab{with ResNet50 backbone using} 
the experimental protocol in~\cite{hendrycks2019anomalyseg}. 
BDD Anomaly 
is a subset of BDD dataset, composed of 6688 street scenes for the training set and 361 for the testing set. The training set {contains} 17 classes, and the test 
\ab{set} is composed of the 17 training classes and 2 OOD classes.
Results in Table~\ref{table:OODresults2} show again that the performances of LDU are close to the ones of Deep Ensembles.

\subsection{Monocular depth experiments}\label{sec:monocular_exp}

We set up our experiments on KITTI dataset~\cite{Uhrig2017THREEDV} with Eigen split training and validation set~\cite{NIPS2014_7bccfde7} to evaluate and compare the predicted depth accuracy and uncertainty quality. We train BTS~\cite{lee2019big} with DenseNet161~\cite{huang2017densely}, and we use the default training setting of BTS (number of epochs, weight decay, batch size) to train DNNs for all uncertainty estimation techniques applied on this backbone.

By default, the BTS baseline does not output uncertainty.  Similarly to 
\ab{\cite{kendall2017uncertainties,ilg2018uncertainty},} we can consider that a DNN may be constructed to find and output the parameters of a parametric distribution (e.g., the mean and variance for a Gaussian distribution). Such networks can be optimized by maximizing their log-likelihood. We denote the result as single predictive uncertainty (Single-PU).
We also train a Deep Ensembles~\cite{lakshminarayanan2017simple} by ensembling 3 DNNs, as well as a MC-Dropout~\cite{gal2016dropout} with eight forward passes.
Without the extra DNNs or training procedures, we also applied Infer-noise~\cite{mi2019training}, which injects Gaussian noise layers to the trained BTS baseline model and propagate eight times to predict the uncertainty.

We have also implemented LDU with the BTS model, but 
we note however that, in the monocular depth estimation setting and in agreement with previous works~\cite{dijk2019neural}, the definition of OOD is fundamentally different with respect to the tasks introduced in the prior experiments. Thus, our objective is to investigate whether LDU is robust, can improve the prediction accuracy and still perform well for aleatoric uncertainty estimation. Table~\ref{table:depth_kitti} lists the depth and uncertainty estimation results on KITTI dataset. \Xuanlong{Using different settings of $\# \vp$ and $\lambda$, the proposed LDU is virtually aligned with the current state-of-the-art, while being significantly lighter computationally (see also Table.~\ref{table:runtime}). More ablation results on the influence of $\# \vp$ and $\lambda$ can be found in the supplementary materials.}

\section{Discussions and Conclusions}
\label{sec:time}

\parag{Discussions.} 
\ab{In Table~\ref{table:runtime} we compare the computational cost of LDU and related methods. For each approach we measure the training (forward+backward) and inference time per image on a NVIDIA RTX 3090Ti and report the corresponding number of parameters.}  
\ab{We report training and inference wall-clock timings averaged over 100 training and validation.} 
We use the same backbones as mentioned in Section~\ref{sec:segment_exp} and Section~\ref{sec:monocular_exp} for semantic segmentation and monocular depth estimation respectively. 
\ab{We note that the runtime of LDU is almost the same as that of the baseline model (standard single forward model). This underpins the efficiency of our approach during inference, a particularly important requirement for practical applications.}

\begin{table}[t]
\begin{center}
\scalebox{0.50}
{
\begin{tabular}{lccclccc} 
\toprule
\multirow{2}{*}{Method} & \multicolumn{3}{c}{\textbf{Semantic segmentation}} & \multicolumn{1}{c}{} & \multicolumn{3}{c}{\textbf{Monocular depth}} \\ 
\cmidrule{2-4}\cmidrule{6-8}
 & Runtime (ms) & Training time (ms) & \#param. &  & Runtime (ms) & Training time (ms) & \#param. \\ 
\toprule
Baseline & \first \textbf{14} &\first \textbf{166.4} & \first\textbf{39.76} &  & \first \textbf{45} & \second 92.8 & \first\textbf{47.00} \\
Deep Ensembles~\cite{lakshminarayanan2017simple}  & \second 56 & 499.2 & \second 119.28 &  & \second 133 & 287.8 & 141.03 \\
MC-Dropout~\cite{gal2016dropout} & 199 & \first \textbf{166.4} & \first\textbf{39.76} &  & 370 &\first \textbf{92.3} & \first\textbf{47.00} \\
Single-PU~\cite{kendall2017uncertainties} & - & - & - &  & \first\textbf{45} & 95.6 & \second 47.01 \\
LDU (ours) & \first \textbf{14} & \second 177.8 & \first\textbf{39.76} &  & \first\textbf{45} & 104.0 & \first\textbf{47.00} \\
\bottomrule
\end{tabular}
}
\vspace{1em}
\caption{
\ab{\textbf{Comparative results for training (forward+backward) and inference wall-clock timings and number of parameters for evaluated methods.} Timings are computed per image and averaged over 100 images.}
}
\vspace{-3em}
\label{table:runtime}
\end{center}
\end{table}
\parag{Conclusions.} \Gianni{In this work, we propose a simple way to modify a DNN 
\ab{to better estimate its predictive uncertainty.}
These minimal changes consist in optimizing \ab{a set of latent} prototypes to learn \ab{to} quantify the uncertainty \ab{by analyzing the position of an input sample in this prototype space}. We perform extensive experiments and show that LDU can outperform state-of-the-art DUMs in most tasks and 
\ab{reach} results comparable to Deep Ensembles with a significant advantage in terms of \ab{computational efficiency and memory requirements.} 
}

\Gianni{Along with the current state of the art methods, 
\ab{a} limitation of our proposed LDU is 
that despite the empirical improvements in uncertainty quantification, it does not provide theoretical guarantees on the correctness of the predicted uncertainty. 
Our perspectives 
\ab{concern further exploration and improvements of} the regularization strategies introduced in LDU on the latent feature representation that would allow us to bound the model error while still 
\ab{preserving} its main task \ab{high} performance.}

\small
\Xuanlong{
\subsection*{Acknowledgement} This work was performed using HPC resources from GENCI-IDRIS (Grant 2020-AD011011970)
and (Grant 2021-AD011011970R1) and Saclay-IA computing platform.}
\normalsize
\clearpage
\bibliographystyle{splncs04}
\bibliography{egbib}
\clearpage
\title{Latent Discriminant deterministic Uncertainty} %
\subtitle{Supplementary Material}

\titlerunning{Latent Discriminant deterministic Uncertainty}
\author{Gianni Franchi\inst{1\dagger} \and
Xuanlong Yu\inst{1,2\dagger} \and
Andrei Bursuc\inst{3} \and
Emanuel Aldea\inst{2} \and
Severine Dubuisson\inst{4} \and
David Filliat\inst{1}}
\authorrunning{Franchi et al.}
\institute{U2IS, ENSTA Paris, Institut polytechnique de Paris \and
SATIE, Paris-Saclay University \and
valeo.ai \and
CNRS, LIS, Aix Marseille University}
\maketitle
\appendix

\addtocontents{toc}{\protect\setcounter{tocdepth}{2}}
{
\hypersetup{linkcolor=black}
\tableofcontents
}

\renewcommand{\theequation}{A\arabic{equation}}
\renewcommand{\thetable}{A\arabic{table}}
\renewcommand{\thefigure}{A\arabic{figure}}

\appendix

\section{Connection with kernel methods}

\Gianni{This section provides a connection between LDU (section 3.2) and kernel ridge regression.}
Let us consider \comEmi{the generic training of a} kernel ridge regressor model~\cite{smola1998learning} $\texttt{f}$ on {the} training set $\{ (\vx_i,y_i)\}_{i=1}^n$, where $n$ is the number of training samples. Let us denote {by} $k$ the kernel.
The \Sev{Representer Theorem} \cite{smola1998learning} states that the solution to this problem is of the form $\texttt{f}(\vx)=\sum_{i=1}^n\alpha_i k(\vx_i,\vx)$, where $\alpha_i$ are the parameters to optimize. 

Therefore, one should evaluate the kernel centered in each training sample. \comEmi{If one relates the proposed LDU model to kernel ridge regression, the main distinction is that} we do not evaluate it across the entire training set, but across a smaller set composed of the prototypes. Hence
$\texttt{f}(\vx)=\sum_{i=1}^m\alpha_i k\left(p_i,h_{\vomega}(\vx_i)\right)$ with $k\left(\cdot,\cdot\right)= \mbox{exp}(-S_c(\cdot,\cdot) )$, which is a composition of kernel positive definite functions. 
\ab{In contrast with} DUQ~\cite{van2020uncertainty} and similarly to SNGP~\cite{liu2020simpleSNGP}
, we approximate the kernel ridge regression based on the prototype set which allows us to simplify the training. 
\ab{However for SNGP, the authors} choose to approximate the kernel differently, by relying on the random projection trick \cite{rahimi2007random}.

\section{Implementation details}

\begin{table}[t]
\renewcommand{\figurename}{Table}
\begin{center}
\scalebox{0.85}
{
\begin{tabular}{l|c}
\toprule
 {\textbf{Hyper-parameter}} &   \textbf{CIFAR-10}   \\ 
\midrule
backbone         &ResNet18   \\ 
\midrule
 initial learning rate         &0.1 \\ 
 \midrule
batch size        &128  \\ 
 \midrule
lr decay ratio     &0.1\\ 
 \midrule
 lr decay epochs         &[80, 160, 200] \\ 
 \midrule
number of train epochs  &250\\ 
\midrule
 weight decay   &0.0001  \\ 
 \midrule
 cutout         & False \\ 
 \midrule
 SyncEnsemble BN        & False\\ 

\bottomrule
\end{tabular}
}
\end{center}
\caption{
{\textbf{Hyper-parameter configuration used in the classification experiments (\S4.1).}}
} 
\label{table:tab2}
\vspace{-3pt}
\end{table}

\begin{table}[h!]
\begin{center}
\scalebox{0.85}
{
\begin{tabular}{l|c|c|c}
\toprule
  \ab{\textbf{Hyper-parameter}} &   \textbf{MUAD} & \textbf{Cityscapes}    & \textbf{BDD-Anomaly}     \\ 
\midrule
Architecture         &Deeplab v3+ & Deeplab v3+ & PSPNet  \\ 
\midrule
backbone         &ResNet50 & ResNet50 & ResNet50  \\ 
\midrule
output stride       &8 &8 &None\\ 
\midrule
learning rate         &0.1 &0.1 & 0.02 \\ 
 \midrule
batch size        &16 & 16 & 4  \\ 
 \midrule
number of train epochs  & 50 & 50 & 30  \\ 
  \midrule
nb Prototypes        & 22 & 22  & 30 \\ 
\midrule
 weight decay  &0.0001 & 0.0001& 0.0001 \\ 
 \midrule
 SyncEnsemble BN        & False & False  & False \\ 
  \midrule
Cutout        & False & False  & False \\ 
  \midrule
 random crop of training images    & 768 & 768 & None \\ 
\bottomrule
\end{tabular}
}
\end{center}
\caption{
{\textbf{Hyper-parameter configuration used in the  semantic segmentation experiments (\S4.2).}}
}\label{table:tab3}
\end{table}

\subsection{Classification and \comEmi{s}emantic segmentation experiments}

This section provides the hyper-parameters used in the classification and semantic-segmentation experiments. Our code is implemented in PyTorch~\cite{paszke2019pytorch}. We used 30 prototypes for BDD-Anomaly and 22 for the other semantic segmentation dataset. $g_{\omega}^{\mbox{unc}}$ is for all the semantic segmentation dataset an MLP with one hidden layer, and the number of neurons in the hidden layer is equal to the number of prototypes. For classification, $g_{\omega}^{\mbox{unc}}$  is a single fully connected layer. All the parameters are introduced in Table \ref{table:tab2} for the classification and Table \ref{table:tab3} for the semantic segmentation.

\subsection{Monocular depth experiments}
As mentioned in the main paper, we train all models using 
the same 
\ab{hyper-parameters} as in the original BTS code~\cite{lee2019big}. We use 30 prototypes in the DM layer for our LDU model. For Single-PU, we duplicate the top layer and double the number of output channels of the pre-logit layer. 
\ab{For the last layer,} we have two one-channel-map outputs: one for depth estimation, one for uncertainty estimation. 
\ab{For the Deep Ensembles baseline, we train Single-PU models.}
\ab{We provide all our hyper-parameters in Table \ref{table:tab4}. We will make the code publicly available after the anonymity period.}

\begin{table}[t!]
\renewcommand{\figurename}{Table}
\begin{center}
\scalebox{0.85}
{
\begin{tabular}{l|c} 
\toprule
\textbf{Hyper-parameter} & \textbf{KITTI} \\ 
\midrule
Architecture & BTS~\cite{lee2019big} \\ 
\midrule
backbone & DenseNet161~\cite{huang2017densely} \\ 
\midrule
initial learning rate & 0.0001 \\ 
\midrule
batch size & 4 \\ 
\midrule
number of train epochs & 50 \\ 
\midrule
weight decay & 0.01 \\ 
\midrule
random crop of training images & (352, 704) \\ 
\midrule
nb Prototypes & 30 \\
\bottomrule
\end{tabular}
}
\end{center}
\caption{
{\textbf{Hyper-parameter configuration used in the monocular depth \ab{estimation} experiments (\S4.3).}}
} 
\label{table:tab4}
\vspace{-3pt}
\end{table}

\section{\Xuanlong{Ablation study}}
In this section, we provide various ablation studies on evaluating the sensitivity of hyper-parameter choices for classification and regression tasks. 

\ab{In Table \ref{table:cifar_lambda} and Table \ref{table:depth_kitti_lambda} we report the results for different values of $\lambda$, the weight of the extra losses for classification on CIRFAR-10 and for monocular depth estimation on KITTI respectively.} 
The performance across metrics is relatively stable with respect to the choice of $\lambda$ with a good compromise at 0.1. 

Concerning the influence of prototype number, according to Table 2 in the main paper and Table \ref{table:depth_kitti_lambda}, the predictive performance is higher with more prototypes, while fewer prototypes make lighter and faster models.

\ab{We also study} the impact of different losses in the classification task in Table~\ref{table:cifar_losses}. The three losses bring consistent improvements \ab{individually and together}.

\begin{table}[h]
\centering
    \begin{tabular}{l|cccc}%
    \toprule
    $\lambda$                  & Acc $\uparrow$   & AUC $\uparrow$  & AUPR $\uparrow$     & ECE $\downarrow$ \\
    \toprule
    0.01            & 87.93 & 0.8425 & 0.9079 & 0.5319  \\
    0.1            & 87.95	&  \textbf{0.8721} &  \textbf{0.9147} &\textbf{ 0.4933}\\
    0.5            & 87.99 & 0.8526 & 0.9109 & 0.5184\\ 
    1.0            & 87.88 & 0.8399 & 0.9035 & 0.5005\\ 
    2.0            & \textbf{88.07} & 0.8339 & 0.9008 & 0.4954 \\
    \bottomrule
    \end{tabular}
    \vspace{1em}
\caption{
      \textbf{Ablation studies for image classification on CIFAR-10.} Sensitivity of $\lambda$ values.
    }
\label{table:cifar_lambda}
\end{table}

\begin{table}[h]
\centering
\begin{tabular}{ccc|cccc}%
    \toprule
     $\mathcal{L}^{\mbox{Unc}}$  &  $\mathcal{L}^{\mbox{Entrop}}$  & $\mathcal{L}^{\mbox{Dis}}$   & Acc $\uparrow$   & AUC $\uparrow$  & AUPR $\uparrow$     & ECE $\downarrow$ \\
    \toprule
    \checkmark & &  &  87.69 & 0.8195 & 0.8600 & 0.5256  \\
     \checkmark & \checkmark &  &  87.87 & 0.8362 &  0.9090 & 0.5386\\
     \checkmark & & \checkmark &  \textbf{88.04} & 0.8613 & 0.8897 & 0.4973\\
     \checkmark & \checkmark & \checkmark &  87.95	&  \textbf{0.8721} &  \textbf{0.9147} & \textbf{0.4933}\\
    \bottomrule
    \end{tabular}
    \vspace{1em}
\caption{
      \textbf{Ablation studies for image classification on CIFAR-10.} Impact of proposed losses ($\lambda{=}0.1$).
    }
\label{table:cifar_losses}
\end{table}

\begin{table}[h]
\begin{center}
\scalebox{0.7}{
\begin{tabular}{rl|cccccccclcc} 
\toprule
 $\# \vp$ & $\lambda$ & d1$\uparrow$ & d2$\uparrow$ & d3$\uparrow$ & Abs Rel$\downarrow$ & Sq Rel$\downarrow$ & RMSE$\downarrow$ & RMSE log$\downarrow$ & log10$\downarrow$ &  & \begin{tabular}[c]{@{}c@{}}AUSE\\RMSE$\downarrow$\end{tabular} & \begin{tabular}[c]{@{}c@{}}AUSE\\Absrel$\downarrow$\end{tabular} \\ 
\toprule
\multirow{4}{*}{30} & 0.01 & \textbf{0.957} & \textbf{0.993} & 0.998 & \textbf{0.061} & \textbf{0.246} & 2.768 & \textbf{0.097} & \textbf{0.027} &  & 0.12 & 0.29 \\
 & 0.1 & 0.955 & 0.992 & 0.998 & \textbf{0.061} & 0.248 & \textbf{2.757} & \textbf{0.097} & \textbf{0.027} &  & 0.09 & 0.26 \\
 & 0.5 & 0.952 & \textbf{0.993} & 0.998 & 0.064 & 0.256 & 2.789 & 0.100 & 0.028 &  & 0.09 & 0.28 \\
 & 1.0 & 0.952 & 0.992 & 0.998 & 0.064 & 0.257 & 2.767 & 0.099 & 0.028 &  & \textbf{0.08} & 0.23 \\ 
\midrule
\multirow{4}{*}{5} & 0.01 & 0.953 & \textbf{0.993} & 0.998 & 0.064 & 0.264 & 2.777 & 0.099 & 0.028 &  & 0.15 & 0.39 \\
 & 0.1 & 0.953 & 0.992 & 0.998 & 0.064 & 0.256 & 2.773 & 0.100 & 0.028 &  & 0.12 & 0.33 \\
 & 0.5 & 0.953 & \textbf{0.993} & 0.998 & 0.063 & 0.253 & 2.776 & 0.099 & 0.028 &  & 0.09 & 0.26 \\
 & 1.0 & 0.954 & \textbf{0.993} & 0.998 & 0.063 & 0.253 & 2.768 & 0.098 & \textbf{0.027} &  & \textbf{0.08} & \textbf{0.21} \\
 \midrule
 15 & 0.1 & 0.954 & \textbf{0.993} & 0.998 & 0.062 & 0.249 & 2.769 & 0.098 & \textbf{0.027} &  & 0.10 & 0.28 \\ 
\bottomrule
\end{tabular}}
\vspace{1em}
\caption{
\textbf{Ablation studies for monocular depth estimation on KITTI.} Sensitivity of $\lambda$ and the number of prototypes.
}
\label{table:depth_kitti_lambda}
\end{center}
\end{table}

\section{\Xuanlong{Training dynamics}}
\begin{figure}[h]
\centering
\begin{center}
\begin{tabular}{ccc}
\includegraphics[width=0.35\linewidth]{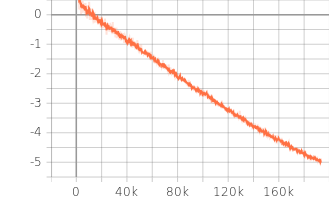} &
\includegraphics[width=0.35\linewidth]{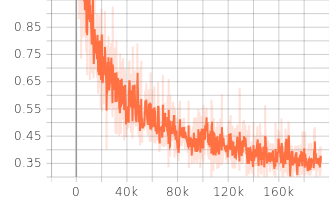} &
\includegraphics[width=0.35\linewidth]{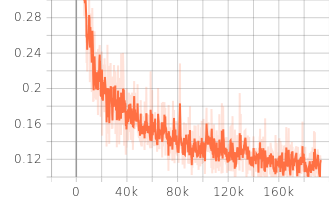} \\
a) $\mathcal{L}^{\mbox{total}}$ (LDU) & b) $\mathcal{L}^{\mbox{Task}}$ (LDU) & c) $\mathcal{L}^{\mbox{Unc}}$ (LDU)\\\\
\includegraphics[width=0.35\linewidth]{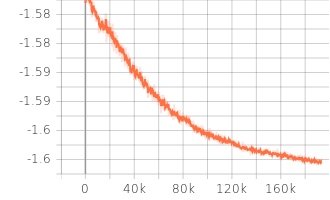} &
\includegraphics[width=0.35\linewidth]{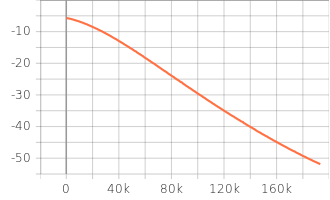} &
\includegraphics[width=0.35\linewidth]{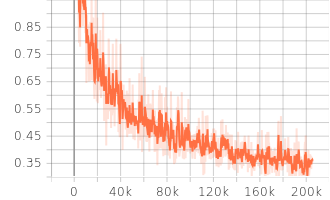}\\
d) $\mathcal{L}^{\mbox{Entrop}}$ (LDU) & e) $\mathcal{L}^{\mbox{Dis}}$ (LDU) & f) $\mathcal{L}^{\mbox{Task}}$ (original)\\
\end{tabular}
\end{center}
\caption{Illustrations on training curves for different losses. Figure a) - e): training curves for the model with LDU modifications; Figure f): training curve for the original model. The $\mathcal{L}^{\mbox{Task}}$ is silog loss for depth regression as defined in BTS~\cite{lee2019big}.}
\label{fig:loss_curves1}
\end{figure}

In this section we illustrate the training curves when we train our models with/without applying LDU modifications. We take our regression experiment as an example, 
\ab{and plot the training curves in Fig.~\ref{fig:loss_curves1}.}
Compared to the original setting, the inserted DM layers and additional losses did not affect the stability of training, all losses decrease smoothly.

\section{More visualizations}
\begin{figure}[t]
\centering
\begin{tabular}{llll}
\rotr{\begin{tabular}[c]{@{}c@{}}\tiny{Image}\end{tabular}} & \includegraphics[width=0.30\linewidth]{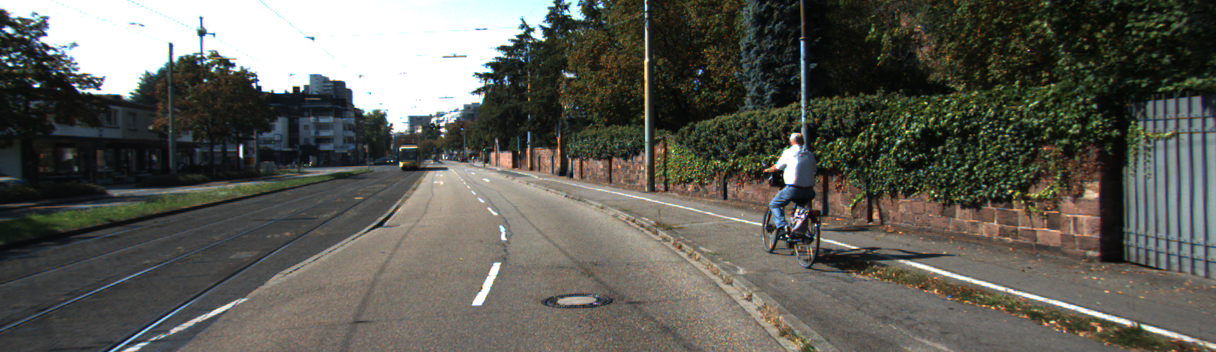}& \includegraphics[width=0.30\linewidth]{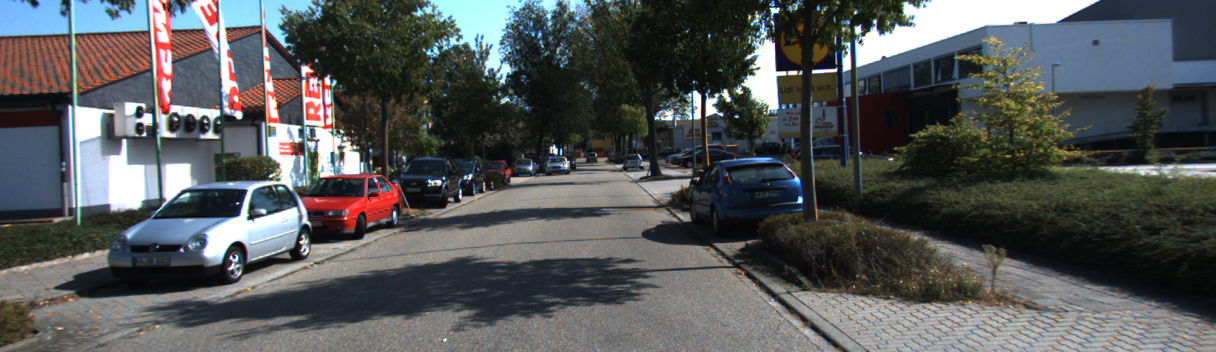} & \includegraphics[width=0.30\linewidth]{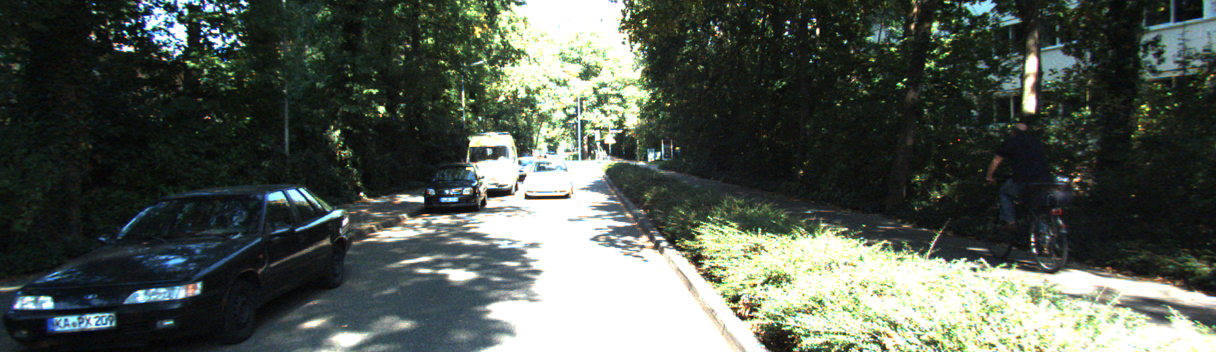} \\
\rotr{\begin{tabular}[c]{@{}c@{}}\tiny{Ground}\\\tiny{truth}\end{tabular}} & \includegraphics[width=0.30\linewidth]{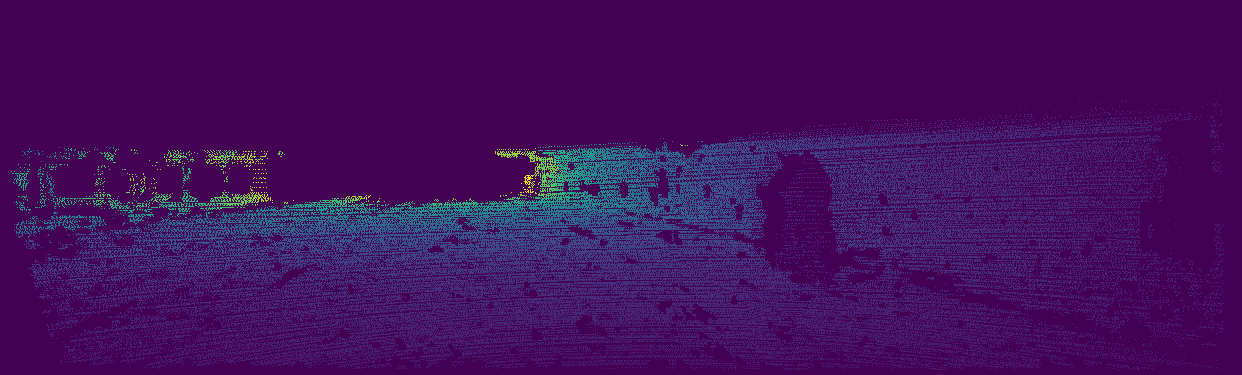} & \includegraphics[width=0.30\linewidth]{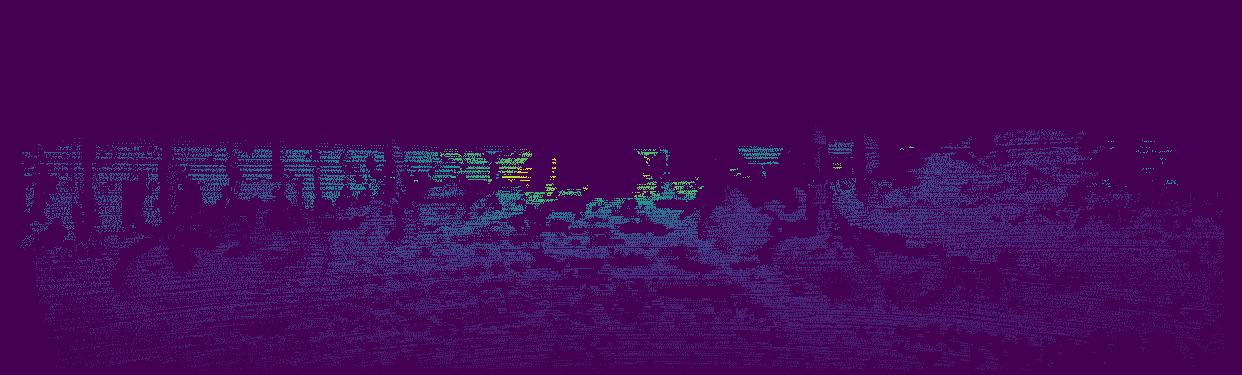} & \includegraphics[width=0.30\linewidth]{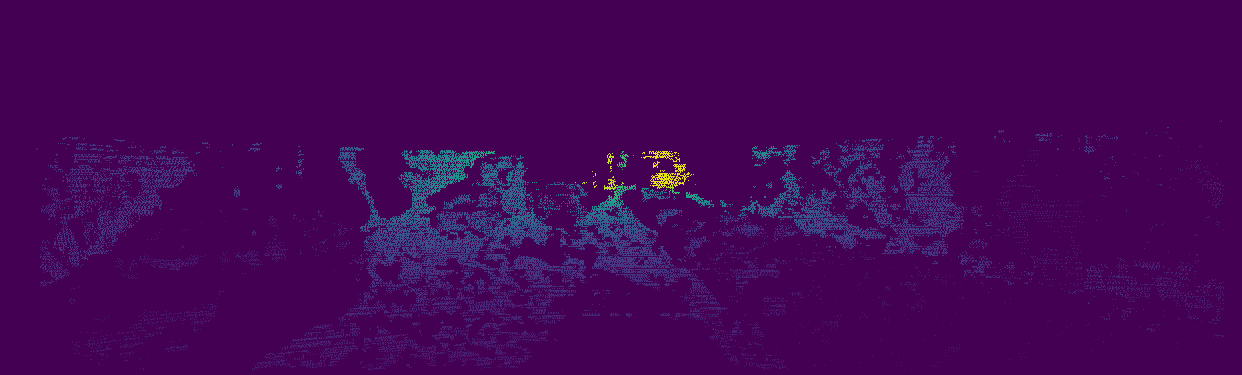} \\
\rotr{\begin{tabular}[c]{@{}c@{}}\tiny{Depth}\\\tiny{prediction}\end{tabular}} & \includegraphics[width=0.30\linewidth]{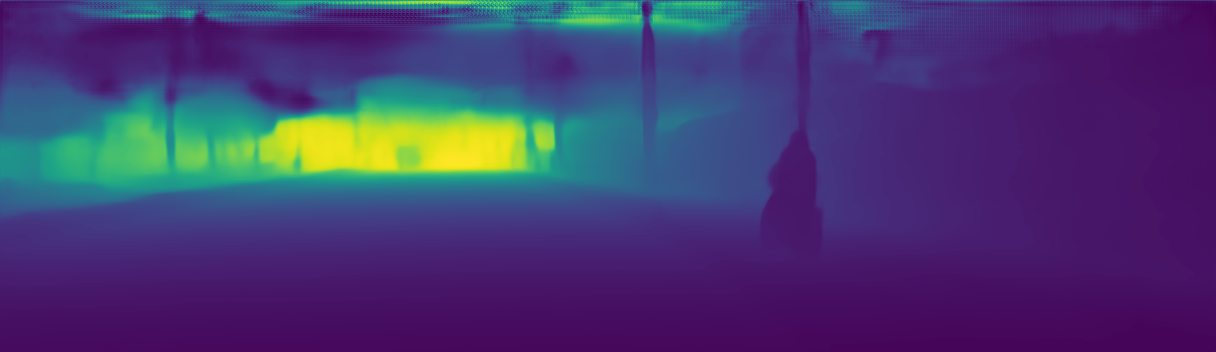} & \includegraphics[width=0.30\linewidth]{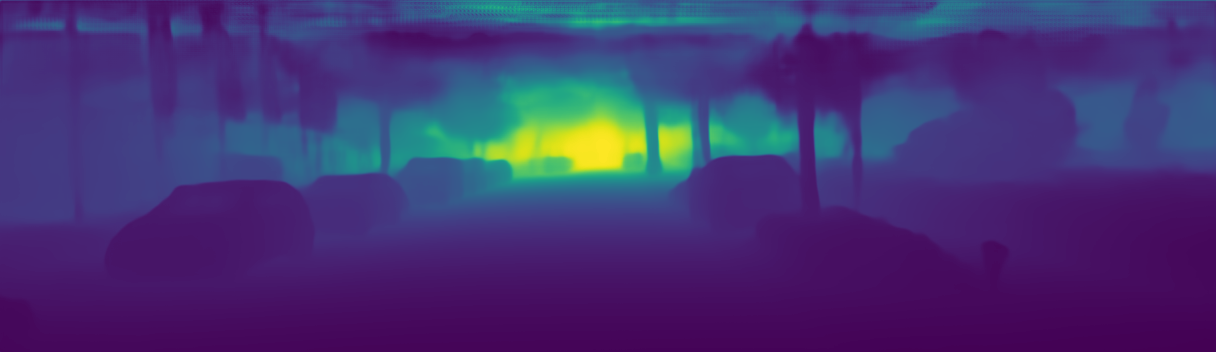} & \includegraphics[width=0.30\linewidth]{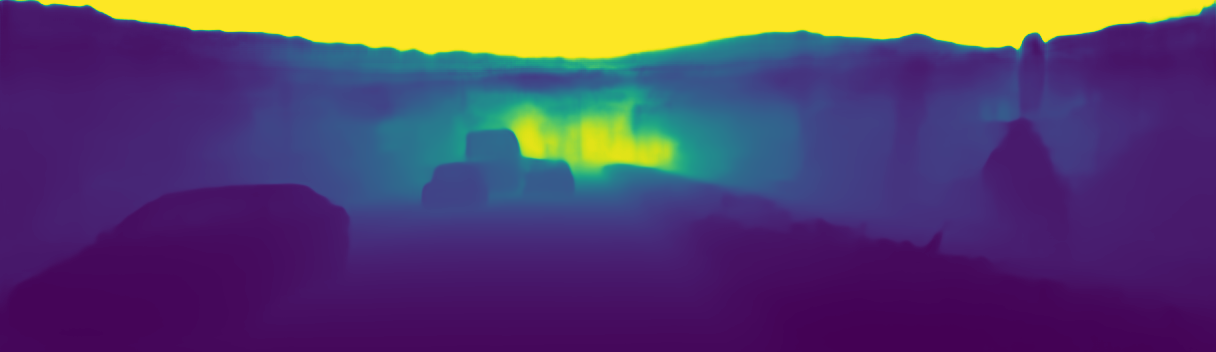} \\
\rotr{\begin{tabular}[c]{@{}c@{}}\tiny{Single-PU}\\\tiny{uncertainty}\end{tabular}} & \includegraphics[width=0.30\linewidth]{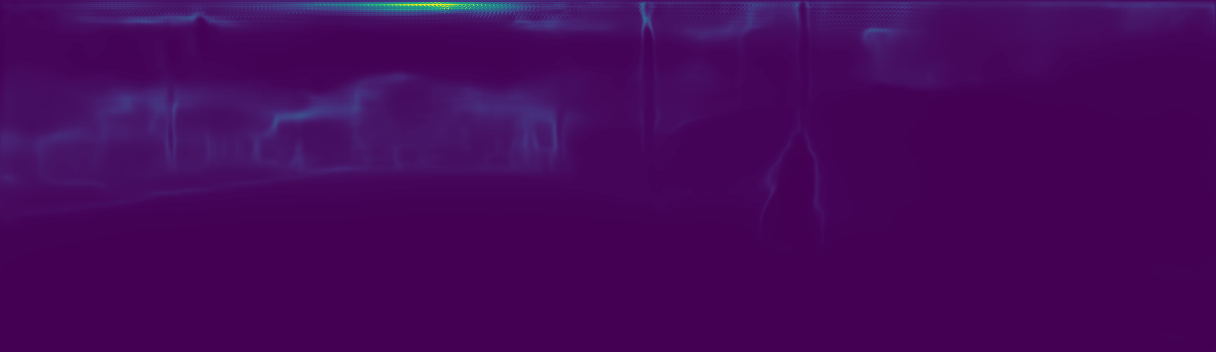}  & \includegraphics[width=0.30\linewidth]{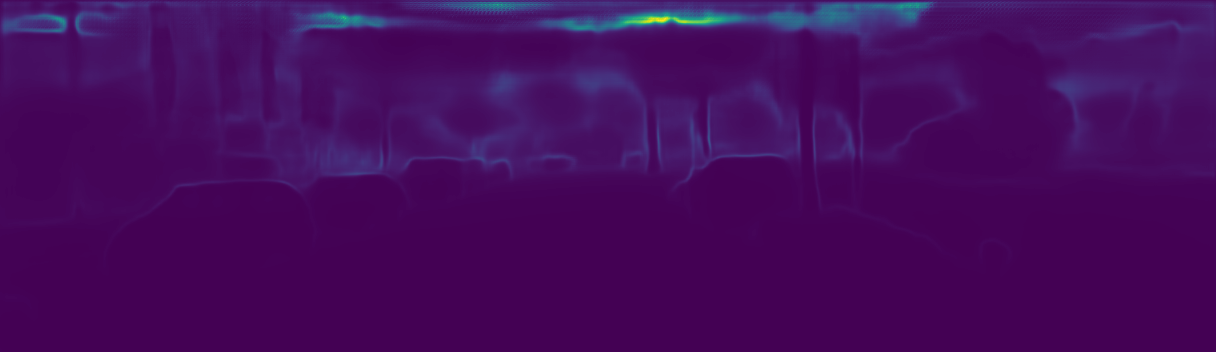}  & \includegraphics[width=0.30\linewidth]{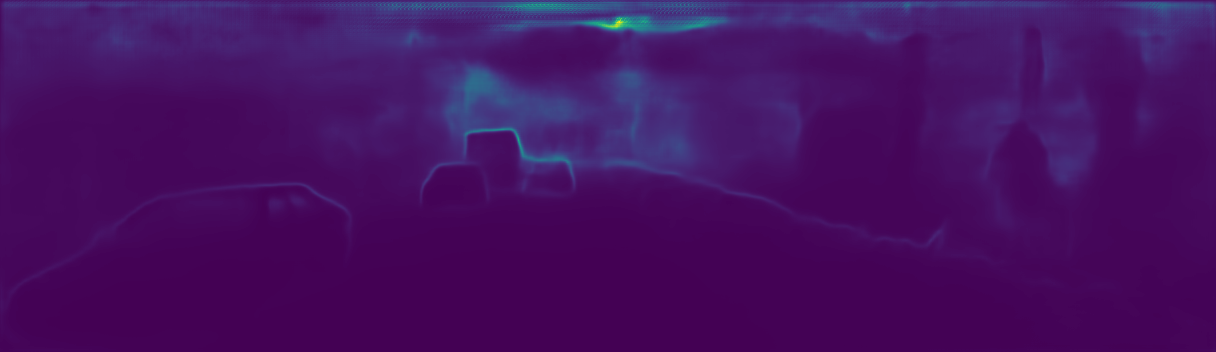}  \\
\rotr{\begin{tabular}[c]{@{}c@{}}\tiny{Deep}\\\tiny{Ensembles}\\\tiny{uncertainty}\end{tabular}} & \includegraphics[width=0.30\linewidth]{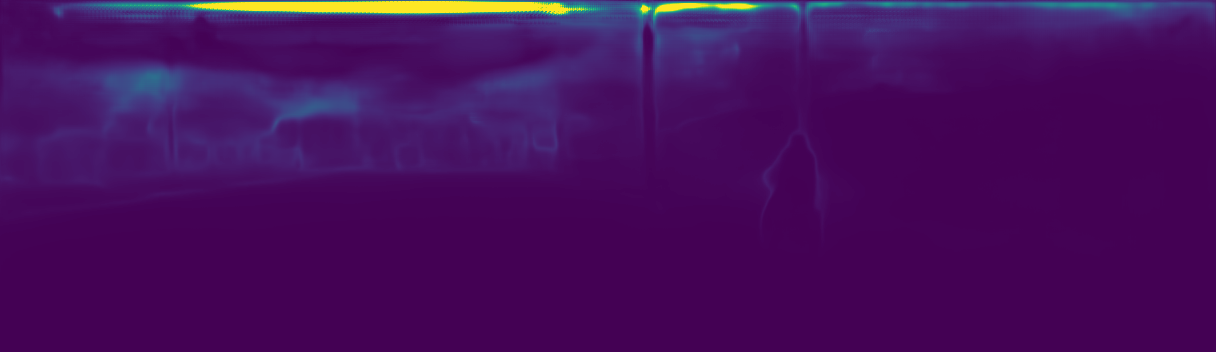} & \includegraphics[width=0.30\linewidth]{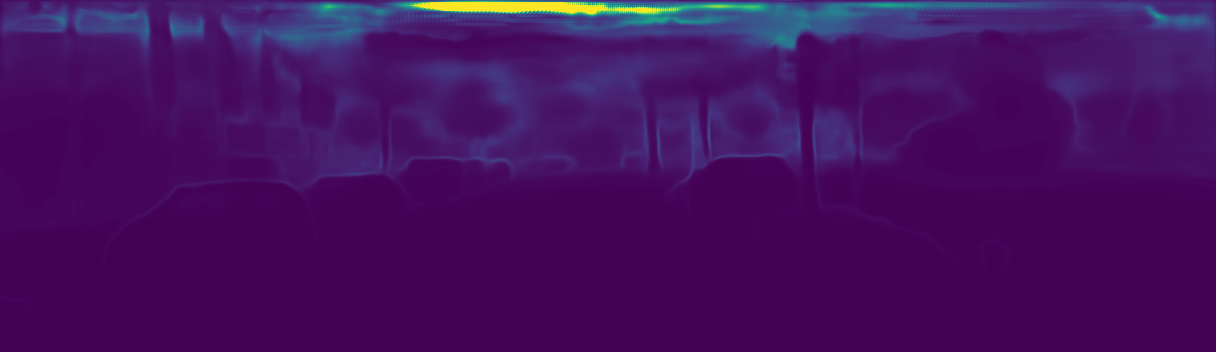} & \includegraphics[width=0.30\linewidth]{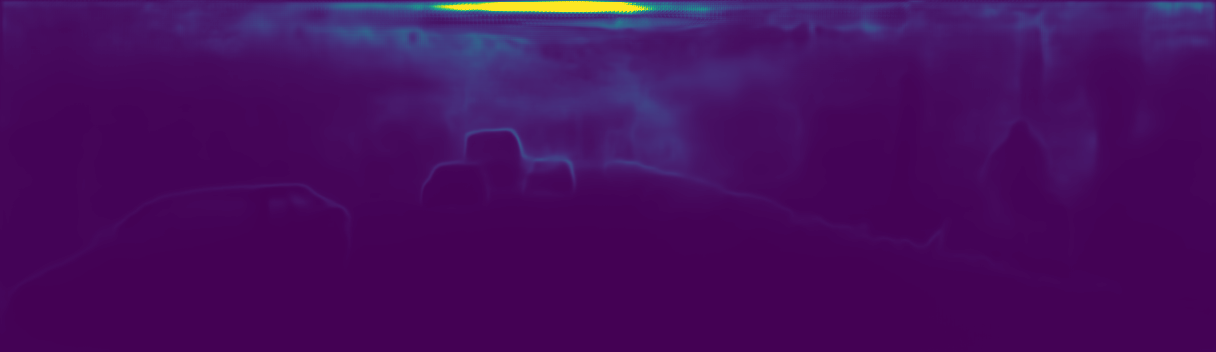} \\
\rotr{\begin{tabular}[c]{@{}c@{}}\tiny{LDU}\\\tiny{uncertainty}\end{tabular}} & \includegraphics[width=0.30\linewidth]{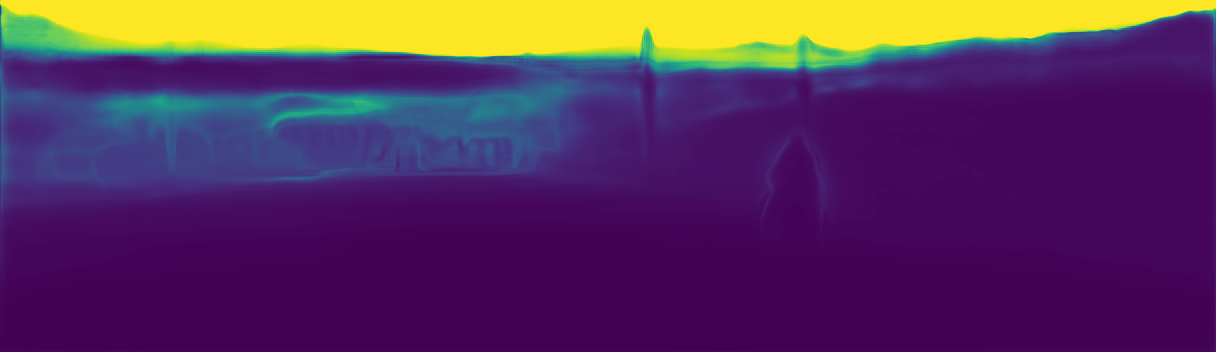} & \includegraphics[width=0.30\linewidth]{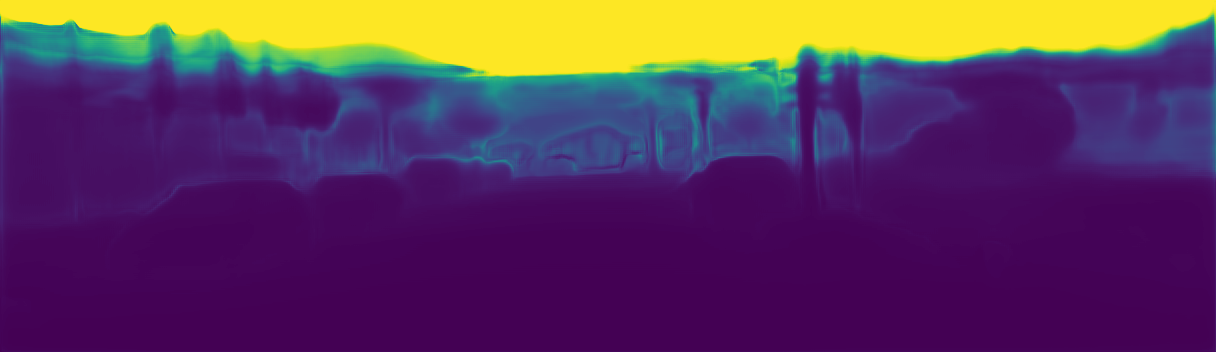} & \includegraphics[width=0.30\linewidth]{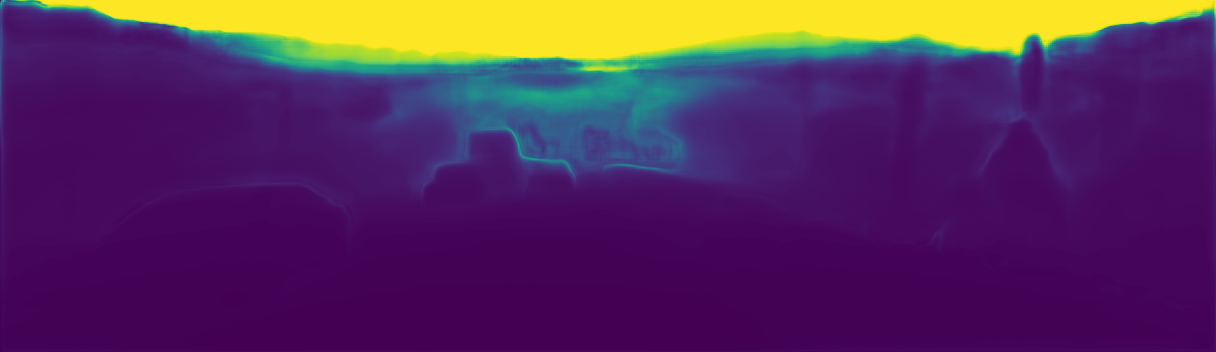}
\end{tabular}
\caption{\ab{Illustration of different uncertainty maps (LDU, Deep Ensembles, Single-PU) on  KITTI images for the monocular depth estimation task~(\S4.3). For both depth and uncertainty maps, the brighter the color is, the bigger the value the pixel has.}
}
\label{fig:monodepth}
\end{figure}

\ab{In Fig.~\ref{fig:monodepth} we present qualitative results depicting uncertainty for monocular depth estimation. We show side-by-side depth predictions and uncertainty maps generated by Single-PU, Deep Ensembles and LDU respectively. We observe that for the areas with valid ground truth, all uncertainty estimation strategies highlight the edges of the objects, where the aleatoric uncertainty is frequently prominent. Concerning the areas without valid ground truth, Deep Ensembles does a better job than Single-PU in highlighting them  since it can capture more epistemic uncertainty due to the ensembling of multiple predictions from the individual models. Our proposed LDU highlights even better some distant areas, especially the upper part of the image where LiDAR beams do not go. We consider that this result stems from LDU regarding this region as OOD regions after training on the entire dataset. }
\end{document}